\documentclass{article} 
\usepackage{masa_arxiv,times}
\usepackage{array}
\usepackage{enumitem}
\usepackage{microtype}
\usepackage{amsmath,amsfonts,bm}

\def\eqref#1{equation~\ref{#1}}

\def\1{\bm{1}}

\def\mA{{\bm{A}}}

\def\mC{{\bm{C}}}
\def\mD{{\bm{D}}}

\def\mH{{\bm{H}}}
\def\mI{{\bm{I}}}

\def\mK{{\bm{K}}}

\def\mM{{\bm{M}}}

\def\mO{{\bm{O}}}

\def\mQ{{\bm{Q}}}
\def\mR{{\bm{R}}}

\def\mV{{\bm{V}}}

\def\mX{{\bm{X}}}

\def\mZ{{\bm{Z}}}

\DeclareMathAlphabet{\mathsfit}{\encodingdefault}{\sfdefault}{m}{sl}
\SetMathAlphabet{\mathsfit}{bold}{\encodingdefault}{\sfdefault}{bx}{n}

\usepackage{hyperref}
\usepackage{url}
\usepackage{graphicx}
\usepackage{amsmath}
\usepackage{wrapfig}
\usepackage{placeins}
\usepackage{hhline}

\usepackage{booktabs}
\usepackage{multirow}
\usepackage[table]{xcolor}
\usepackage{makecell}

\definecolor{BaseRow}{HTML}{F7FAFD}
\definecolor{MASArow}{HTML}{EEF6FF}
\definecolor{AvgCol}{HTML}{F7F3E8}

\newcommand{\masa}{MASA}
\newcommand{\dash}{--}
\newcommand{\best}[1]{\textbf{#1}}
\newcommand{\reported}{\textsuperscript{\ensuremath{\dagger}}}

\newcommand{\basecell}[1]{\cellcolor{BaseRow}#1}
\newcommand{\masacell}[1]{\cellcolor{MASArow}#1}
\newcommand{\avgcell}[1]{\cellcolor{AvgCol}#1}

\newcolumntype{C}[1]{>{\centering\arraybackslash}m{#1}}
\newcolumntype{L}[1]{>{\raggedright\arraybackslash}m{#1}}

\definecolor{GainText}{HTML}{087A2E}

\renewcommand{\eqref}[1]{\textup{Eq.~(\ref{#1})}}

\definecolor{BestCol}{HTML}{EEF6FF}

\newcommand{\benchlabel}[1]{\rotatebox[origin=c]{90}{\scriptsize\textit{#1}}}

\title{Sparse Attention Is Matrix Approximation, \\Not Choosing from a Bag of Values}

\author{
Fang Wan$^{1}$,
Xufeng Liu$^{1}$,
Fan Li$^{2}$,
Yi Liu$^{1}$\thanks{Correspondence to yi.liu.4@stonybrook.edu} \\
$^{1}$ Stony Brook University \quad $^{2}$ Duke University
}

\begin{document}

\maketitle
\vspace{-6mm}
\begin{abstract}
\vspace{-3mm}
Large Language Models (LLMs) achieve strong performance across many domains, but their efficiency is limited by the quadratic cost of attention with respect to prompt length. Sparse attention reduces this cost by retaining only a small fraction of query-key interactions to approximate the full attention matrix. However, existing methods are trapped in a mathematically wrong view: they simply keep large scalar entries or high-mass regions of the attention matrix. This treats the attention matrix as a bag of values, ignoring that it is used as a structured matrix whose entries jointly determine the attention output through multiplication with value vectors. We argue that this is the core conceptual issue: \textbf{sparse attention should be formulated as matrix approximation, not as blindly choosing the largest values from a bag of entries.} Based on this view, we propose \underline{\textbf{M}}atrix \underline{\textbf{A}}pproximation \underline{\textbf{S}}parse \underline{\textbf{A}}ttention (\underline{\textbf{MASA}}). MASA replaces raw attention-mass ranking with a closed-form score that measures how much each sparse unit reduces matrix-product approximation error. As a theory-grounded plug-in correction, MASA can be added to existing sparse attention frameworks without changing their sparse kernels or budgets. Extensive experiments across multiple sparse attention methods, benchmarks, and LLM backbones show consistent accuracy gains, supporting both MASA and the matrix-approximation view of sparse attention.
\end{abstract}
\vspace{-2mm}
\section{Introduction}
\vspace{-2mm}
Large Language Models (LLMs) have achieved strong performance across a wide range of applications, from natural language processing~\citep{radford2018improving} to mathematical reasoning~\citep{ahn2024large}, multimodal tasks~\citep{team2023gemini}, and scientific discovery~\citep{zhang2025artificial}. However, efficiency remains a central challenge for LLMs~\citep{bai2024beyond}. As tasks become more complex and prompts become more structured, the input length continues to grow rapidly~\citep{liu2025world}. Since attention computation has quadratic complexity with respect to the prompt length~\citep{vaswani2017attention}, attention becomes a major bottleneck for efficient long-context inference.

Sparse attention is a natural way to reduce this cost while keeping the main modeling benefit of attention. Prior work has observed that attention matrices often contain sparse and structured patterns~\citep{jiang2024minference}, suggesting that only a small fraction of query-key interactions is needed to approximate full attention. Based on this observation, many methods select a sparse subset of the attention matrix under a compute budget. Early approaches use fixed or predefined patterns: Sliding Window Attention~\citep{jiang2023mistral7b} restricts each query to a local context window, while MInference~\citep{jiang2024minference} approximates attention using vertical lines, slash lines, and blocks. Other methods make the selection input-aware: SeerAttention~\citep{gao2024seerattention} learns to predict block-sparse attention patterns, and Quest~\citep{tang2024quest} retrieves query-aware key pages. More recent methods further adapt the sparse budget, such as FlexPrefill~\citep{laiflexprefill}, which decides when to stop selecting additional sparse units, and Tactic~\citep{zhu2025tactic}, which adjusts the budget through clustering and distribution fitting. \textbf{Despite these different designs, the final selection rule remains ultimately magnitude-driven: sparse attention methods keep large scalar entries or high-mass regions of the attention matrix.}

\begin{wrapfigure}{r}{0.55\textwidth}
    \vspace{-4mm}
    \centering
    \includegraphics[width=\linewidth]{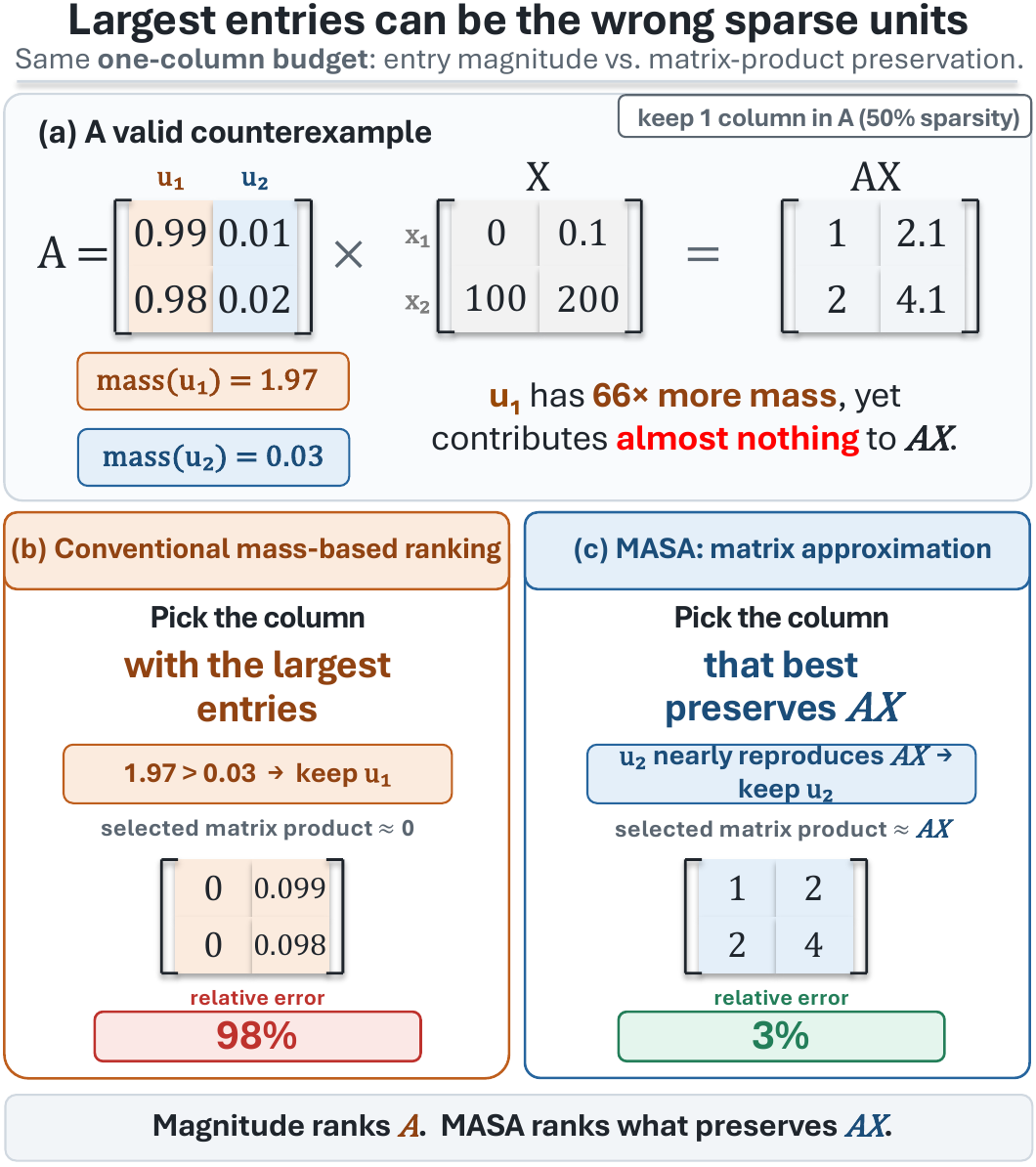}
    \vspace{-8mm}
    \caption{ A minimal two-column counterexample. Under the same one-column budget,
conventional mass-based ranking selects $u_1$, whose mass is $66\times$ larger but
incurs $98\%$ relative matrix-product error. MASA selects $u_2$ and reduces
the error to $3\%$. Here, $\mX$ is an illustrative probe matrix.
}
    \label{fig:masa_example}
    \vspace{-15pt}
\end{wrapfigure}

This selection principle has a mathematical mismatch. As illustrated in
Fig.~\ref{fig:masa_example}, a candidate with much larger mass can
preserve the matrix product substantially worse under the same sparse budget.
Magnitude-based selection treats the attention matrix as a bag of scalar values,
whereas Transformers use it as a structured matrix acting on value vectors
through matrix multiplication~\citep{vaswani2017attention}. Therefore, the
right question is not which entries of the attention matrix are largest, but
which selected entries best preserve the matrix product that produces the
attention output. In other words, \textit{sparse attention should be formulated as
matrix approximation, not as blindly choosing the largest values from a bag of
entries.}

Based on this view, we propose \underline{\textbf{M}}atrix \underline{\textbf{A}}pproximation \underline{\textbf{S}}parse \underline{\textbf{A}}ttention (\underline{\textbf{MASA}}), a theory-grounded plug-in correction for sparse attention. MASA keeps the sparse units, kernels, and budgets of existing sparse attention methods, but replaces their raw attention-mass ranking with a closed-form score derived from matrix-product approximation error. This changes the selection target from preserving large entries of the attention matrix to preserving the matrix action that determines the attention output. As a result, MASA can be added to diverse sparse attention frameworks without changing their implementation interface.

In this work, our contributions are summarized as follows: 
\vspace{-1mm}
\begin{itemize}[
    leftmargin=1em,
    labelsep=0.45em,
    topsep=2pt,
    itemsep=1.5pt,
    parsep=0pt,
    partopsep=0pt
]
    \item \textbf{Theory.}
    We identify a long-standing selection mismatch in sparse attention: existing magnitude-based rules approximate the attention matrix entrywise, while the Transformer uses attention through matrix multiplication. We provide a matrix-approximation view that gives the correct selection objective. 

    \item \textbf{Method.}
    We propose MASA, a theory-grounded plug-in scoring method that ranks sparse units by their closed-form reduction in matrix-product approximation error. MASA can be integrated into existing sparse attention baselines while keeping their sparse kernels and budgets unchanged.

    \item \textbf{Experiments.}
    We evaluate MASA across multiple sparse attention methods, benchmarks, and LLM backbones. Under matched settings with the original baselines, MASA gives consistent accuracy gains, supporting both the proposed method and the matrix-approximation view of sparse attention.
\end{itemize}
\vspace{-2mm}
\section{Preliminaries}
\vspace{-3mm}

We review sparse attention methods through three representative frameworks used to evaluate MASA, then formalize sparse attention. Broader LLM acceleration methods are discussed in Appendix~\ref{app:llm_acceleration}.

\vspace{-3mm}
\subsection{Related Work}
\vspace{-2mm}
\label{sec:baselines}

MInference~\citep{jiang2024minference} is a training-free framework that exploits \textbf{multiple structured patterns through offline head-wise assignment}. It searches among A-shaped, vertical-slash, and block-sparse patterns for each attention head. The assigned pattern type and budget remain fixed, while sparse indices are constructed dynamically for each input. MoA~\citep{fu2025mixture} similarly uses offline configuration, but sets head- and layer-specific sliding-window spans.

SeerAttention~\citep{gao2024seerattention} instead \textbf{learns input-dependent sparse layouts within a block-based representation}. GPU-friendly blocks can approximate different attention patterns without explicit head-wise pattern-type assignment. A lightweight gate learns block importance from dense attention supervision, providing a representative learned selector under the fixed budget used in our evaluation. SpargeAttn~\citep{zhangspargeattention} provides a training-free alternative using online attention prediction and softmax-aware filtering.

FlexPrefill~\citep{laiflexprefill} further increases flexibility by \textbf{adapting both pattern type and budget without additional training}. For each input and head, it chooses between vertical-slash and query-aware block selection, then determines the number of retained units using a cumulative-attention threshold. XAttention~\citep{xuxattention} also performs input-dependent block selection, estimating block importance through antidiagonal sampling.

Together, MInference, SeerAttention, and FlexPrefill cover the main design choices in sparse attention: vertical, slash, and block units; offline-assigned and input-adaptive pattern types; fixed and adaptive budgets; and training-free and learned selection. Their complementary mechanisms provide a suitable testbed for evaluating \textit{whether MASA gives a general scoring correction across distinct sparse attention designs, rather than an improvement confined to a single implementation}.
\vspace{-2mm}
\subsection{Formalization}
\vspace{-2mm}
In attention, $\mQ$, $\mK$, and $\mV$ are the query, key, and value matrices, where
$\mQ,\mK,\mV\in\mathbb{R}^{N\times d}$. Their height is the prompt sequence length
$N$, which can grow very large and become the main bottleneck in attention
computation. 
After applying the causal mask, the full causal attention matrix and output are
\[
\mA
=
\operatorname{softmax}\left(
\frac{\mQ\mK^{\top}}{\sqrt{d}}
+
\mM
\right),
\qquad
\mO
=
\mA\mV,
\qquad
\mM_{ij}
=
\begin{cases}
0, & j \le i, \\
-\infty, & j > i.
\end{cases}
\]
Here $\mA\in\mathbb{R}^{N\times N}$ is the full causal attention matrix.

However, full causal attention still computes attention scores for all valid
query-key pairs, leading to $\mathcal{O}(N^2)$ complexity. Sparse attention reduces
this cost by retaining only a selected subset of valid query-key pairs. Let
$\mathcal{S}$ denote the sparse attention index set, and let
$\mM_{\mathcal{S}}\in\{0,1\}^{N\times N}$ be its binary retention mask defined by
$\mM_{\mathcal{S}}[i,j]=\mathbf{1}[(i,j)\in\mathcal{S}]$.

Since $\mathcal{S}$ only contains valid causal positions, the causal constraint is
already included in the definition of $\mathcal{S}$. For selection analysis, we write the sparse attention contribution as
\[
\mO_{\mathcal{S}}
=
(\mM_{\mathcal{S}}\odot\mA)\mV .
\]
This notation isolates the retained support $\mathcal{S}$. In actual sparse-attention implementations, the selected support is passed to the baseline kernel, which may apply its own sparse-softmax normalization or other kernel-specific details. MASA does not modify this sparse computation; it only changes how $\mathcal{S}$ is ranked before the baseline kernel is called.

Normal causal attention is the special case where
$\mathcal{S}=\{(i,j)\mid 1 \le j \le i \le N\}$.

\vspace{-2mm}
\section{Method}
\vspace{-2mm}
We first identify the selection mismatch shared by existing sparse attention methods in Sec.~\ref{sec:issue}. We then introduce MASA as a matrix-approximation correction in Sec.~\ref{sec:masa}, and discuss its implementation in the sparse prefilling setting in Sec.~\ref{sec:masa_practical}.
\vspace{-2mm}
\subsection{Issue}
\vspace{-2mm}
\label{sec:issue}

The goal of sparse attention is to reduce the number of selected query-key pairs while
keeping the sparse attention output close to the full attention output. This can be
written as a two-objective problem~\citep{laiflexprefill}:
\begin{equation}
\label{eq:objective}
\min_{\mathcal{S}}
\left\|
\mA\mV
-
(\mM_{\mathcal{S}}\odot \mA)\mV
\right\|_F^2,
\qquad
\min_{\mathcal{S}}
|\mathcal{S}|.
\end{equation}
The first objective measures the output error between full and sparse attention, and the second objective measures the sparse attention cost. Here $\|\cdot\|_F^2$ is the squared Frobenius norm, which sums squared errors over all entries of a matrix; see Appendix~\ref{app:frobenius_norm}. We use this form because it gives a closed-form MASA score, while more general matrix-action metrics are discussed in
Appendix~\ref{app:masa_metric}.

Under a fixed compute budget $B$, the sparse selection problem in~\eqref{eq:objective}
becomes
\begin{equation}
\label{eq:objective_budget}
\mathcal{S}^{\star}
=
\arg\min_{\mathcal{S}}
\left\|
\mA\mV
-
(\mM_{\mathcal{S}}\odot \mA)\mV
\right\|_F^2
\quad
\mathrm{s.t.}
\quad
|\mathcal{S}|\le B .
\end{equation}
This is the objective implied by attention computation: the selected sparse pattern
should preserve the matrix product $\mA\mV$.

\textbf{Current sparse selection.}
Most existing sparse attention methods do not directly estimate the output error in~\eqref{eq:objective_budget}. In sparse prefill implementations, they first define candidate sparse units, such as blocks, vertical lines, slash lines, or query-aware index groups, as discussed in Sec.~\ref{sec:baselines}. For a candidate unit $u$, they assign a scalar entrywise score
\begin{equation}
\label{eq:scalar_assign}
s_{\mathrm{entry}}(u)
=
\sum_{(i,j)\in u}
\phi(\mA_{ij}),
\end{equation}
where $\phi$ maps each attention entry to a scalar score. In most cases, $\phi$
corresponds to attention mass, but it can also be maximum mass, thresholded mass, or
a learned gate-score variant~\citep{gao2024seerattention}. The sparse pattern is
then formed by keeping the units with the largest $s_{\mathrm{entry}}(u)$ under the
budget. This rule reduces the attention matrix to entry scores $\mA\longmapsto
\{\phi(\mA_{ij})\}_{i,j}$.
The corresponding entrywise approximation loss is
\[
\label{eq:entrywise_loss}
\left\|
\mA
-
\mM_{\mathcal{S}}\odot\mA
\right\|_F^2
=
\sum_{(i,j)\notin\mathcal{S}}
\mA_{ij}^{2}.
\]
Since $\sum_{i,j}\mA_{ij}^{2}$ is fixed, minimizing this loss under a fixed budget is
equivalent to keeping the largest squared attention entries:
\begin{equation}
\label{eq:entrywise_topk}
\arg\min_{|\mathcal{S}|\le B}
\left\|
\mA
-
\mM_{\mathcal{S}}\odot\mA
\right\|_F^2
=
\arg\max_{|\mathcal{S}|\le B}
\sum_{(i,j)\in\mathcal{S}}
\mA_{ij}^{2}.
\end{equation}
Therefore, current sparse selection asks which entries or regions of
$\mA$ are large. It does not ask which selected entries best preserve the actual
attention output $\mA\mV$.

\textbf{Mismatch.}
The Transformer does not consume the attention matrix as isolated scalar entries.
It uses the attention matrix as a matrix acting on value vectors. From~\eqref{eq:objective_budget}, the true output error is 
\[
\label{eq:true_output_error}
\left\|
\mA\mV
-
(\mM_{\mathcal{S}}\odot \mA)\mV
\right\|_F^2
=
\left\|
\big(\mA-\mM_{\mathcal{S}}\odot\mA\big)\mV
\right\|_F^2 .
\]
Expanding this matrix-product error gives
\begin{equation}
\label{eq:expansion}
\left\|
\big(\mA-\mM_{\mathcal{S}}\odot\mA\big)\mV
\right\|_F^2
=
\operatorname{tr}
\left(
\big(\mA-\mM_{\mathcal{S}}\odot\mA\big)
\mV\mV^{\top}
\big(\mA-\mM_{\mathcal{S}}\odot\mA\big)^{\top}
\right).
\end{equation}
This shows that \textbf{entrywise magnitude ranking is correct only when $\mV\mV^{\top}$ behaves like an identity matrix} over token positions. This condition
does not hold in long-context attention. Since $\mV\in\mathbb{R}^{N\times d}$ and
usually $N\gg d$,
\[
\operatorname{rank}(\mV\mV^{\top})
\le d \ll N .
\]
Therefore, $\mV\mV^{\top}$ cannot be an identity matrix over the $N$ token positions.

This is the core mismatch. Existing sparse attention methods mainly answer:
\emph{which attention entries are large?}
The matrix-approximation problem asks a different question:
\emph{which selected entries best preserve $\mA\mV$?}
These two questions are not equivalent.
\vspace{-2mm}
\subsection{MASA}
\label{sec:masa}
\vspace{-2mm}
\textbf{Main idea.}
The objective in~\eqref{eq:objective_budget} shows that sparse attention should
preserve the matrix product $\mA\mV$, rather than only large entries of $\mA$.
We propose \underline{\textbf{M}}atrix \underline{\textbf{A}}pproximation
\underline{\textbf{S}}parse \underline{\textbf{A}}ttention
(\underline{\textbf{MASA}}), a plug-in scoring correction for sparse attention.
MASA keeps the same candidate sparse units and the same budget as the baseline,
such as blocks, vertical lines, slash lines, or query-aware index groups. It only
replaces the score used to rank these units.

Let $\mX\in\mathbb{R}^{N\times r}$ be a probe matrix. The dense probe output and
sparse probe output are
\[
\label{eq:probe_outputs}
\mO_{\mX}
=
\mA\mX,
\qquad
\mO_{\mX,\mathcal{S}}
=
(\mM_{\mathcal{S}}\odot\mA)\mX .
\]
MASA scores sparse patterns by the matrix-product approximation error
\begin{equation}
\label{eq:masa_action_objective}
\mathcal{L}_{\mathrm{MASA}}(\mathcal{S};\mX)
=
\left\|
\mO_{\mX}
-
\mO_{\mX,\mathcal{S}}
\right\|_F^2
=
\left\|
(\mA-\mM_{\mathcal{S}}\odot\mA)\mX
\right\|_F^2 .
\end{equation}
When $\mX=\mV$,~\eqref{eq:masa_action_objective} becomes exactly the attention
output error in~\eqref{eq:objective_budget}. Thus, MASA changes sparse selection
from preserving large entries of $\mA$ to preserving the matrix product
$\mA\mX$. Fig.~\ref{fig:masa_example} gives a minimal two-column instance of this
objective: although $u_2$ has much lower mass, it receives a higher
MASA score because it preserves $\mA\mX$ substantially more accurately. \textbf{We use the general notation $\mX$ rather than $\mV$ because the probe can also be pooled, norm-controlled, or projected value information, as discussed in
Sec.~\ref{sec:masa_practical}.}

Expanding~\eqref{eq:masa_action_objective} gives
\[
\label{eq:masa_action_trace}
\mathcal{L}_{\mathrm{MASA}}(\mathcal{S};\mX)
=
\operatorname{tr}
\left(
(\mA-\mM_{\mathcal{S}}\odot\mA)
\mX\mX^{\top}
(\mA-\mM_{\mathcal{S}}\odot\mA)^{\top}
\right).
\]
This has the same form as~\eqref{eq:expansion}, with $\mX\mX^{\top}$ replacing
$\mV\mV^{\top}$. Therefore, entrywise magnitude ranking appears only as the
identity-Gram special case.

\textbf{Reduction score.}
Directly optimizing~\eqref{eq:masa_action_objective} over all sparse patterns is
combinatorial, so sparse attention methods usually rank candidate units one by one.
MASA keeps this ranking interface. For a candidate sparse unit $u$, let
$\mM_u\in\{0,1\}^{N\times N}$ be its binary retention mask. Its contribution to the
probe output is $\mC_u=(\mM_u\odot\mA)\mX$.
MASA scores $u$ by the reduction in squared matrix-product error obtained by retaining this contribution:
\begin{equation}
\label{eq:masa_unit_score}
s_{\mathrm{MASA}}(u)
=
\|\mO_{\mX}\|_F^2
-
\|\mO_{\mX}-\mC_u\|_F^2
=
2\langle \mO_{\mX},\mC_u\rangle_F
-
\|\mC_u\|_F^2 .
\end{equation}
The sparse pattern is formed by keeping the units with the largest $s_{\mathrm{MASA}}(u)$ under the same budget as the baseline. If the budget is adaptive, units with non-positive scores can simply be skipped. Compared with the entrywise score in~\eqref{eq:scalar_assign}, \textbf{MASA asks whether a unit helps reconstruct $\mA\mX$, not whether the entries inside the unit are large.}

For a single query-key pair $(i,j)$, let $\mO_{\mX,i}$ and $\mX_j$ denote the
corresponding row vectors of $\mO_{\mX}$ and $\mX$. Then
\begin{equation}
\label{eq:masa_pair_score}
s_{\mathrm{MASA}}(i,j)
=
2\mA_{ij}
\left\langle
\mO_{\mX,i},\mX_j
\right\rangle
-
\mA_{ij}^{2}
\left\|
\mX_j
\right\|_2^2 .
\end{equation}
This closed form shows why attention magnitude alone is insufficient. The score also depends on the direction and norm of the probe vector $\mX_j$, and on its alignment with the dense matrix action $\mO_{\mX,i}$. 

The connection to entrywise ranking follows directly from~\eqref{eq:masa_pair_score}.
If $\mX\mX^{\top}=\mI$, then
\[
\langle\mO_{\mX,i},\mX_j\rangle=\mA_{ij},\qquad
\|\mX_j\|_2^2=1,\qquad
s_{\mathrm{MASA}}(i,j)=\mA_{ij}^{2}.
\]
Thus, squared-magnitude ranking is only the identity-Gram special case of MASA.
As discussed in Sec.~\ref{sec:issue}, this condition does not hold for
value-derived probes in long-context attention.

\vspace{-2mm}
\subsection{Implementation}
\label{sec:masa_practical}
\vspace{-2mm}
Our theory applies to sparse attention selection in general. In our implementation and experiments, we instantiate MASA in the sparse prefill setting, where the main cost comes from computing attention over a long prompt before decoding. This is also the setting of our baselines, including MInference~\citep{jiang2024minference}, SeerAttention~\citep{gao2024seerattention}, and FlexPrefill~\citep{laiflexprefill}. MASA keeps their sparse unit definitions, sparse kernels, and budget rules unchanged, and only replaces the score used to rank candidate sparse units. 

All MASA scores are computed at the same profile resolution as the baselines. Thus, $\mA$ and $\mX$ in the scoring formulas denote the baseline's profile attention map and the matching profile probe when the baseline ranks units at block or other reduced resolution. MASA therefore does not require a new full-resolution dense attention map beyond the information used for sparse-unit ranking.

\textbf{Structured sparse units.}
MASA can be used with the same sparse unit family as the baseline. For any candidate unit $u$, the exact score is the reduction score in~\eqref{eq:masa_unit_score}. When a unit contains at most one key position per query row, this score reduces to the sum of pair scores:
\[
\label{eq:masa_unit_sum}
s_{\mathrm{MASA}}(u)
=
\sum_{(i,j)\in u}
s_{\mathrm{MASA}}(i,j).
\]
This applies to vertical and slash units~\citep{jiang2024minference,laiflexprefill}:
\[
s_{\mathrm{MASA}}(j)
=
\sum_i s_{\mathrm{MASA}}(i,j),
\qquad
s_{\mathrm{MASA}}(\ell)
=
\sum_{i-j=\ell}
s_{\mathrm{MASA}}(i,j).
\]

For block units, the full score in~\eqref{eq:masa_unit_score} should be used because multiple key positions in the same query row interact through the squared norm term. If $I$ and $J$ are the query and key index ranges of a block, the block score is
\begin{equation}
\label{eq:masa_block_score}
s_{\mathrm{MASA}}(I,J)
=
\sum_{i\in I}
\left[
2
\left\langle
\mO_{\mX,i},
\sum_{j\in J}\mA_{ij}\mX_j
\right\rangle
-
\left\|
\sum_{j\in J}\mA_{ij}\mX_j
\right\|_2^2
\right].
\end{equation}
The second term keeps the interaction among entries inside the same block, so the block is scored by matrix-product error reduction rather than by entry magnitude.

\textbf{Memory-safe computation.}
A direct implementation of~\eqref{eq:masa_pair_score} may seem to require all vectors $\mA_{ij}\mX_j$, which would form an $N\times N\times r$ tensor. MASA avoids this tensor. First compute
\[
\label{eq:masa_memory_precompute}
\mO_{\mX}
=
\mA\mX,
\qquad
\mC
=
\mO_{\mX}\mX^{\top},
\qquad
\mathbf{n}_j
=
\|\mX_j\|_2^2 .
\]
Then all pair scores are obtained by
\[
\label{eq:masa_memory_safe}
\mD
=
2\mA\odot\mC
-
\mA^{\odot 2}\odot(\mathbf{1}\mathbf{n}^{\top}),
\]
where $\mD_{ij}=s_{\mathrm{MASA}}(i,j)$, $\mA^{\odot 2}$ denotes entrywise square, and $\mathbf{1}$ is the all-one column vector. This computation only uses matrix products and score matrices at the profile resolution, without materializing the $N\times N\times r$ contribution tensor.

For vertical and slash units, we aggregate entries of $\mD$ according to the unit definition of the baseline. For block units, we use~\eqref{eq:masa_block_score} and compute the block contribution $\sum_{j\in J}\mA_{ij}\mX_j$ within the block. This keeps the scoring memory-safe while retaining the block-level reduction score.

\textbf{Probe choice.}
The direct probe is $\mX=\mV$, because the true attention output is $\mA\mV$. Raw value vectors, however, can contain high-norm outliers. Since attention output depends on both attention weights and value vectors~\citep{kobayashi2020attention}, large value norms can dominate the score in~\eqref{eq:masa_pair_score}. Similar outlier effects have also been shown to harm Transformer computation and quantization robustness~\citep{bondarenko2023quantizable}. We therefore use norm-controlled value probes to keep value directions while reducing the effect of extreme norms.

One option is soft norm balancing:
\begin{equation}
\label{eq:masa_v_norm_alpha}
\mX_j
=
\frac{\mV_j}
{(\|\mV_j\|_2+\epsilon)^{\alpha}},
\qquad
0\le \alpha\le 1 .
\end{equation}
When $\alpha=0$, this is the raw value probe. When $\alpha=1$, it keeps the value direction while removing the value norm. Intermediate values keep part of the norm information while reducing the effect of high-norm outliers.

Another option is norm clipping:
\begin{equation}
\label{eq:masa_v_norm_clip}
\mX_j
=
\mV_j
\cdot
\min\left(
1,
\frac{c}{\|\mV_j\|_2+\epsilon}
\right),
\end{equation}
where $c$ is a chosen percentile of value-vector norms in the profile region. This keeps normal value vectors unchanged and only clips unusually large norms.

The probe view also \textbf{supports more general matrix-action metrics}. Appendix~\ref{app:masa_metric} shows that weighted output metrics can be written in the same form by choosing an appropriate probe $\mX$, and that the Frobenius matrix-product error controls the spectral matrix-product error. Finally, when the baseline ranks units on a profile attention map, we build $\mX$ at the same profile resolution. For example, block-level selection uses block-pooled value probes, so that the probe \textbf{matches the sparse units used by the implemented kernel}.
\vspace{-2mm}
\section{Experiments}
\vspace{-2mm}
\label{sec:experiments}
Existing sparse attention studies
often differ in budgets, kernels, LLM backbones, and benchmark choices, which makes isolated comparisons hard to interpret. We therefore evaluate MASA under a controlled protocol designed to test whether the proposed matrix-approximation view improves existing sparse attention methods \textbf{without changing their own settings}. First, we do not evaluate MASA on a single favorable framework, but attach it to three representative sparse attention baselines with different sparse units and selection rules. Second, we do not tune the sparse budget, kernel, baseline hyperparameters, or data processing for MASA; each baseline keeps its official configuration, and MASA only replaces the sparse-unit ranking score. Third, we evaluate on the long-context benchmarks used by the original baselines, including RULER~\citep{hsieh2024ruler}, InfiniteBench~\citep{zhang2024bench}, and LongBench~\citep{bai2024longbench}. This protocol directly tests \textbf{whether MASA improves sparse attention under the conditions where the baselines are already designed to work well}.

\vspace{-2mm}
\subsection{Setups}
\vspace{-2mm}
\label{sec:setups}
\textbf{Baselines.} 
We evaluate MASA by augmenting three representative sparse attention methods: MInference~\citep{jiang2024minference}, SeerAttention~\citep{gao2024seerattention}, and FlexPrefill~\citep{laiflexprefill}, as discussed in Sec.~\ref{sec:baselines}. These baselines cover vertical, slash, and block-wise sparse patterns; fixed and input-adaptive selection; fixed and adaptive budgets; and training-free and learned selection rules. This makes them a strong test bed for evaluating whether MASA is a general correction rather than a method-specific modification. 

\textbf{Implementation.} 
All experiments are conducted on a single NVIDIA H100 80GB GPU with CUDA 12.6. The FlashAttention~\citep{dao2022flashattention} and Triton~\citep{tillet2019triton} settings follow the corresponding baseline implementations. For each sparse attention baseline, MASA uses the official code path, released weights, sparse kernels, sparse unit definitions, and budget rules. The only change is the ranking score used to select candidate sparse units. This matched setting isolates the effect of the MASA score from other engineering factors.

\textbf{Benchmarks.} 
We follow the benchmark choices of the original baselines. RULER~\citep{hsieh2024ruler} is used for all three baselines. It is a synthetic long-context benchmark with controllable sequence lengths and task types, including needle-in-a-haystack retrieval, multi-hop tracing, and aggregation. InfiniteBench~\citep{zhang2024bench} is used for MInference and FlexPrefill. It evaluates long-dependency understanding over contexts with an average of about 214K tokens and includes both synthetic and real-world tasks. LongBench~\citep{bai2024longbench} is used for SeerAttention. It is a bilingual long-context benchmark covering multi-document question answering, summarization, and retrieval tasks. Using the baselines' own benchmark settings avoids choosing tasks that favor MASA.

\textbf{LLM Backbones.}
Following the baseline settings, we use
\href{https://huggingface.co/gradientai/Llama-3-8B-Instruct-262k}
{Llama-3-8B-Instruct-262k}
for MInference,
\href{https://huggingface.co/meta-llama/Llama-3.1-8B-Instruct}
{Llama-3.1-8B-Instruct}
for SeerAttention, and both Llama-3.1-8B-Instruct and
\href{https://huggingface.co/Qwen/Qwen2-7B-Instruct}
{Qwen2-7B-Instruct}
for FlexPrefill. These test MASA across different LLM backbones rather than a single model family.

\definecolor{ResultDeltaPos}{HTML}{17723B}
\definecolor{ResultDeltaNeg}{HTML}{B33636}
\definecolor{ResultDeltaZero}{HTML}{666666}

\newcommand{\resultdeltapos}[1]{\textcolor{ResultDeltaPos}{+#1}}
\newcommand{\resultdeltaneg}[1]{\textcolor{ResultDeltaNeg}{#1}}
\newcommand{\resultdeltazero}{\textcolor{ResultDeltaZero}{0}}

\newcommand{\resultframeworklabel}[1]{%
  \rotatebox[origin=c]{90}{%
    \fontsize{7.4}{7.7}\selectfont\bfseries #1%
  }%
}

\newcommand{\resultbackbonelabel}[1]{%
  \rotatebox[origin=c]{90}{%
    \fontsize{7.2}{7.5}\selectfont\itshape #1%
  }%
}

\newcommand{\resultframeworksep}{%
  \specialrule{0.90pt}{1.4pt}{1.4pt}%
}

\newcommand{\resultfirstsep}{%
  \specialrule{0.90pt}{0pt}{1.4pt}%
}

\newcolumntype{G}[1]{%
  >{\columncolor{AvgCol}\centering\arraybackslash}m{#1}%
}

\newcommand{\rulerreferencerule}{%
  \arrayrulecolor{black!28}%
  \hhline{~~*{7}{-}}
  \arrayrulecolor{black}%
}

\newcommand{\infreferencerule}{%
  \arrayrulecolor{black!28}%
  \hhline{~~*{11}{-}}
  \arrayrulecolor{black}%
}

\newcommand{\rulerbackbonerule}{%
  \arrayrulecolor{black!65}%
  \hhline{~*{8}{-}}\hhline{~~~~~~~~~*{2}{-}}
  \arrayrulecolor{black}%
}

\newcommand{\infbackbonerule}{%
  \arrayrulecolor{black!65}%
  \hhline{~*{12}{-}}\hhline{~~~~~~~~~~~~~*{2}{-}}
  \arrayrulecolor{black}%
}

\definecolor{TblDeltaPos}{HTML}{17723B}
\definecolor{TblDeltaNeg}{HTML}{B33636}
\definecolor{TblDeltaZero}{HTML}{666666}

\newcommand{\tblfwlabel}[1]{%
  \rotatebox[origin=c]{90}{%
    \fontsize{7.4}{7.7}\selectfont\bfseries #1%
  }%
}
\newcommand{\tblbblabel}[1]{%
  \rotatebox[origin=c]{90}{%
    \fontsize{7.2}{7.5}\selectfont\itshape #1%
  }%
}
\newcommand{\tbldpos}[1]{\textcolor{TblDeltaPos}{+#1}}
\newcommand{\tbldneg}[1]{\textcolor{TblDeltaNeg}{#1}}
\newcommand{\tbldzero}{\textcolor{TblDeltaZero}{0}}

\begin{table}[t]
\centering
\vspace{-8mm}
\caption{
RULER~\citep{hsieh2024ruler} results under matched sparse-attention settings. Within each framework, Original and \masa{} use the same sparse implementation and differ only in the ranking rule; Full Attention is the reference. $\Delta_{\mathrm{Full}}$ is the signed Avg. difference from Full Attention.
Higher is better. \textbf{Bold} marks the better Avg. between Original and \masa{}.
}
\label{tab:ruler_main}
\small
\setlength{\tabcolsep}{3.1pt}
\renewcommand{\arraystretch}{1.22}

\begin{tabular}{
@{}
C{0.46cm}
@{\hspace{0.7pt}}
C{0.42cm}
@{\hspace{2\tabcolsep}}
l
c c c c c c
|
C{0.70cm}
C{0.86cm}
}
\toprule
& & Variant
& 4K
& 8K
& 16K
& 32K
& 64K
& 128K
& \avgcell{Avg.}
& \avgcell{$\Delta_{\mathrm{Full}}$} \\

\resultfirstsep

\multirow[c]{3}{0.46cm}{%
  \centering\resultframeworklabel{MInfer}%
}
&
& Full Attention
& 97.2
& 91.8
& 87.3
& 80.8
& 77.4
& 72.2
& \avgcell{84.4}
& \avgcell{\resultdeltazero} \\

\hhline{~~*{9}{-}}
&
& Original
& 97.7
& 91.2
& 88.5
& 85.0
& 82.3
& 77.6
& \avgcell{87.0}
& \avgcell{\resultdeltapos{2.6}} \\

&
& \masacell{\masa}
& \masacell{97.93}
& \masacell{91.29}
& \masacell{88.50}
& \masacell{85.45}
& \masacell{81.57}
& \masacell{78.88}
& \avgcell{\best{87.27}}
& \avgcell{\best{\resultdeltapos{2.87}}} \\

\resultframeworksep

\multirow[c]{3}{0.46cm}{%
  \centering\resultframeworklabel{SeerAttn}%
}
&
& Full Attention
& 95.53
& 92.37
& 92.01
& 87.63
& 84.39
& 76.26
& \avgcell{88.01}
& \avgcell{\resultdeltazero} \\

\hhline{~~*{9}{-}}

&
& Original
& 95.53
& 92.71
& 92.02
& 88.49
& 83.48
& 73.37
& \avgcell{87.60}
& \avgcell{\resultdeltaneg{-0.41}} \\

&
& \masacell{\masa}
& \masacell{95.53}
& \masacell{92.74}
& \masacell{92.84}
& \masacell{89.56}
& \masacell{84.01}
& \masacell{73.85}
& \avgcell{\best{88.09}}
& \avgcell{\best{\resultdeltapos{0.08}}} \\

\resultframeworksep

\multirow[c]{6}{0.46cm}{%
  \centering\resultframeworklabel{FlexPrefill}%
}
&
\multirow[c]{3}{0.42cm}{%
  \centering\resultbackbonelabel{Llama}%
}
& Full Attention
& 95.67
& 93.75
& 93.03
& 87.26
& 84.37
& 78.13
& \avgcell{88.70}
& \avgcell{\resultdeltazero} \\

\hhline{~~*{9}{-}}
&
& Original
& 95.43
& 93.51
& 94.71
& 89.42
& 82.93
& 79.09
& \avgcell{89.18}
& \avgcell{\resultdeltapos{0.48}} \\

&
& \masacell{\masa}
& \masacell{95.43}
& \masacell{93.99}
& \masacell{94.95}
& \masacell{89.66}
& \masacell{85.09}
& \masacell{80.54}
& \avgcell{\best{89.94}}
& \avgcell{\best{\resultdeltapos{1.24}}} \\

\cmidrule[0.58pt]{2-11}

&
\multirow[c]{3}{0.42cm}{%
  \centering\resultbackbonelabel{Qwen}%
}
& Full Attention
& 89.90
& 88.70
& 80.77
& 79.33
& 56.49
& 17.79
& \avgcell{68.83}
& \avgcell{\resultdeltazero} \\

\hhline{~~*{9}{-}}

&
& Original
& 90.39
& 89.91
& 83.17
& 81.25
& 59.14
& 20.67
& \avgcell{70.75}
& \avgcell{\resultdeltapos{1.92}} \\

&
& \masacell{\masa}
& \masacell{90.88}
& \masacell{88.94}
& \masacell{84.85}
& \masacell{82.64}
& \masacell{62.79}
& \masacell{22.83}
& \avgcell{\best{72.16}}
& \avgcell{\best{\resultdeltapos{3.33}}} \\

\bottomrule
\end{tabular}
\vspace{-4mm}
\end{table}
\begin{table*}[t]
\centering
\vspace{-1mm}
\caption{
InfiniteBench~\citep{zhang2024bench} results under matched sparse-attention settings.
Within each framework, Original and \masa{} use the same sparse implementation and differ only in the ranking rule; Full Attention is the reference. $\Delta_{\mathrm{Full}}$ is the signed Avg. difference from Full Attention.
Higher is better. \textbf{Bold} marks the better Avg. between Original and \masa{}.
}
\label{tab:infinitebench_main}
\scriptsize
\setlength{\tabcolsep}{2.35pt}
\renewcommand{\arraystretch}{1.20}

\resizebox{\textwidth}{!}{%
\begin{tabular}{
@{}
C{0.46cm}
@{\hspace{0.7pt}}
C{0.42cm}
@{\hspace{1.5pt}}
l
c c c c c c c c c c
|
C{0.70cm}
C{0.86cm}
@{}
}
\toprule
& & Variant
& En.Sum
& En.QA
& En.MC
& En.Dia
& Zh.QA
& \makecell[c]{Code\\Debug}
& \makecell[c]{Math\\Find}
& \makecell[c]{Retr.\\PassKey}
& \makecell[c]{Retr.\\Number}
& \makecell[c]{Retr.\\KV}
& \avgcell{Avg.}
& \avgcell{$\Delta_{\mathrm{Full}}$} \\

\resultfirstsep

\multirow[c]{3}{0.46cm}{%
  \centering\resultframeworklabel{MInfer}%
}
&
& Full Attention
& 20.2
& 12.4
& 67.3
& 6.0
& 12.9
& 22.1
& 26.6
& 100.0
& 100.0
& 14.4
& \avgcell{38.2}
& \avgcell{\resultdeltazero} \\

\cline{3-15}

&
& Original
& 20.5
& 12.9
& 65.9
& 7.5
& 12.5
& 22.3
& 33.1
& 100.0
& 100.0
& 12.8
& \avgcell{38.8}
& \avgcell{\resultdeltapos{0.6}} \\

&
& \masacell{\masa}
& \masacell{19.73}
& \masacell{12.15}
& \masacell{66.81}
& \masacell{8.50}
& \masacell{12.28}
& \masacell{22.84}
& \masacell{33.43}
& \masacell{100.00}
& \masacell{100.00}
& \masacell{16.20}
& \avgcell{\best{39.20}}
& \avgcell{\best{\resultdeltapos{1.00}}} \\

\resultframeworksep

\multirow[c]{6}{0.46cm}{%
  \centering\resultframeworklabel{FlexPrefill}%
}
&
\multirow[c]{3}{0.42cm}{%
  \centering\resultbackbonelabel{Llama}%
}
& Full Attention
& 31.91
& 25.92
& 69.43
& 21.50
& 31.95
& 16.75
& 24.29
& 99.15
& 99.66
& 60.00
& \avgcell{48.06}
& \avgcell{\resultdeltazero} \\

\cline{3-15}

&
& Original
& 31.82
& 24.82
& 69.43
& 19.50
& 35.46
& 16.75
& 31.14
& 98.64
& 99.83
& 44.00
& \avgcell{47.14}
& \avgcell{\resultdeltaneg{-0.92}} \\

&
& \masacell{\masa}
& \masacell{32.10}
& \masacell{26.25}
& \masacell{66.81}
& \masacell{19.50}
& \masacell{33.66}
& \masacell{20.05}
& \masacell{29.14}
& \masacell{98.31}
& \masacell{99.66}
& \masacell{54.40}
& \avgcell{\best{47.99}}
& \avgcell{\best{\resultdeltaneg{-0.07}}} \\

\cmidrule[0.58pt](lr){2-15}

&
\multirow[c]{3}{0.42cm}{%
  \centering\resultbackbonelabel{Qwen}%
}
& Full Attention
& 4.65
& 5.57
& 34.50
& 9.00
& 11.27
& 24.87
& 24.57
& 95.42
& 75.42
& 0.00
& \avgcell{29.53}
& \avgcell{\resultdeltazero} \\

\cline{3-15}

&
& Original
& 14.27
& 6.55
& 34.06
& 13.50
& 11.51
& 20.81
& 30.86
& 96.95
& 72.20
& 0.00
& \avgcell{30.07}
& \avgcell{\resultdeltapos{0.54}} \\

&
& \masacell{\masa}
& \masacell{15.55}
& \masacell{6.38}
& \masacell{32.31}
& \masacell{16.50}
& \masacell{13.09}
& \masacell{21.07}
& \masacell{30.29}
& \masacell{94.41}
& \masacell{74.58}
& \masacell{0.00}
& \avgcell{\best{30.42}}
& \avgcell{\best{\resultdeltapos{0.89}}} \\

\bottomrule
\end{tabular}%
}
\vspace{-5mm}
\end{table*}

\vspace{-2mm}
\subsection{Main Results}
\vspace{-3mm}
\label{sec:main_results}

Tables~\ref{tab:ruler_main}--\ref{tab:longbench_main} summarize the main results; Appendix~\ref{app:full_tables} provides the full comparisons. Within each framework, Original and MASA use the same LLM backbone, benchmark, candidate units, sparse kernel, and sparse budget, differing only in the ranking score. Full Attention provides the dense reference, and $\Delta_{\mathrm{Full}}$ reports the signed average-score difference from it. Each Avg. summarizes the corresponding task or length set. \textbf{MASA improves all eight matched average scores across three sparse methods, three benchmark suites, and both Llama- and Qwen-family backbones.}

These gains are obtained under the baselines' released configurations, where the original sparse methods already approach or exceed the dense reference.\footnote{When a sparse baseline already approaches full-attention quality, the remaining score deficit is limited. The relevant question is how much a ranking correction recovers or adds at the same sparse budget, rather than whether it produces a large margin over the dense model. Full attention is a reference, not a strict upper bound on downstream accuracy.} \textbf{MASA nevertheless delivers substantial gains relative to the dense-reference differences in several settings.} On InfiniteBench, MASA raises FlexPrefill-Llama from $47.14$ to $47.99$ against a full-attention score of $48.06$, recovering approximately $92\%$ of the original average-score deficit. On RULER, MASA raises SeerAttention from $87.60$ to $88.09$, moving it from below the dense reference to above it. For FlexPrefill-Llama on RULER, replacing the ranking score adds $0.76$ points, exceeding the original method's $0.48$-point advantage over full attention and increasing that advantage to $1.24$ points. Thus, MASA improves \textbf{already strong sparse systems} without increasing their sparse budgets or changing their kernels.

\begin{wraptable}{r}{0.45\textwidth}
\centering
\vspace{-5mm}
\caption{
LongBench~\citep{bai2024longbench} results under matched sparse-attention
settings. Within SeerAttention, Original and \masa{} differ only in the
sparse-unit ranking rule; Full Attention is the reference.
$\Delta_{\mathrm{Full}}$ is the signed Avg. difference from Full Attention.
Higher is better. \textbf{Bold} marks the better Avg. and its
$\Delta_{\mathrm{Full}}$.
}
\label{tab:longbench_main}

\small
\setlength{\tabcolsep}{3.8pt}
\renewcommand{\arraystretch}{1.15}

\resizebox{\linewidth}{!}{%
\begin{tabular}{
@{}
>{\raggedright\arraybackslash}m{1.80cm}
>{\centering\arraybackslash}m{0.72cm}
>{\centering\arraybackslash}m{0.72cm}
>{\centering\arraybackslash}m{0.72cm}
!{{\color{black!50}\vrule width 0.45pt}}
>{\columncolor{AvgCol}\centering\arraybackslash}m{0.72cm}
>{\columncolor{AvgCol}\centering\arraybackslash}m{0.88cm}
@{}
}
\toprule
Variant
& 0--4K
& 4--8K
& 8K+
& Avg.
& $\Delta_{\mathrm{Full}}$ \\
\midrule

Full Attention
& 55.32
& 53.98
& 52.90
& 54.07
& \tbldzero \\

\cmidrule[0.28pt](r){1-4}
\cmidrule[0.28pt](l){5-6}

Original
& 55.43
& 54.49
& 52.69
& 54.20
& \tbldpos{0.13} \\

\masacell{\masa}
& \masacell{55.45}
& \masacell{54.48}
& \masacell{53.16}
& \best{54.37}
& \best{\tbldpos{0.30}} \\

\bottomrule
\end{tabular}%
}
\end{wraptable}

\textbf{The benefits are particularly evident at longer context lengths, the main target of sparse attention acceleration.}
Across the four RULER configurations at 32K, 64K, and 128K, MASA improves 11 of 12 matched length-level scores. For every configuration, the mean gain over these longer contexts at 32K--128K exceeds that at 4K--16K. At 128K, MASA improves MInference, SeerAttention, FlexPrefill-Llama, and FlexPrefill-Qwen by $1.28$, $0.48$, $1.45$, and $2.16$ points, respectively. At 64K, the gains reach $2.16$ points for FlexPrefill-Llama and $3.65$ points for FlexPrefill-Qwen. On LongBench, MASA improves SeerAttention's 8K+ score from $52.69$ to $53.16$. These results support the value of matrix-approximation ranking for long-context sparse prefill under the original frameworks' sparse budgets and kernels.

Beyond downstream accuracy, Appendix~\ref{app:attention_output_error} compares executed sparse-attention outputs with dense attention. The evaluated MASA configurations reduce normalized squared output error in all four framework--backbone pairs, providing direct empirical support for matrix-product preservation.

\vspace{-3mm}
\subsection{Efficiency}
\vspace{-2mm}
\label{sec:efficiency}
MASA replaces the baseline's sparse-unit ranking score rather than adding a new attention component. Its only additional computation is evaluating the reduction score in~\eqref{eq:masa_pair_score} within the existing ranking stage. We therefore measure kernel-level latency under the same sparse settings to verify that this replacement preserves the efficiency advantage of sparse attention.

\begin{figure*}[t]
    \centering
    \vspace{-8mm}
    \includegraphics[width=\textwidth]{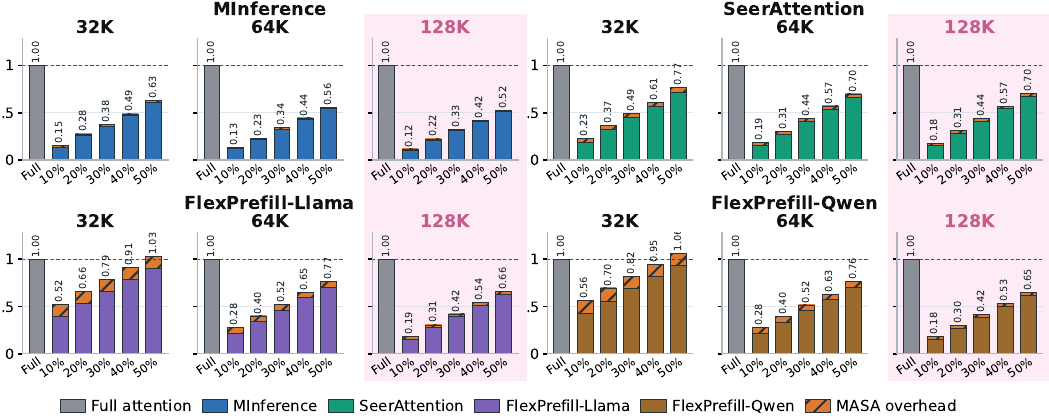}
    \vspace{-20pt}
    \caption{Long-context kernel-level latency normalized to dense FlashAttention-2 at sequence lengths 32K, 64K, and 128K. The x-axis gives the retained sparse budget ratio. All methods use the same latency protocol; method-specific sparse kernels and selection or gating modules follow their official implementations.}
    \label{fig:MASA_latency}
    \vspace{-5mm}
\end{figure*}

Fig.~\ref{fig:MASA_latency} reports latency normalized to dense FlashAttention-2 at sequence lengths 32K, 64K, and 128K. Across MInference, SeerAttention, FlexPrefill-Llama, and FlexPrefill-Qwen, replacing the original ranking score with MASA adds empirically negligible overhead relative to the sparse attention runtime and remains far below the dense attention cost. The overhead also stays negligible at 128K, the main regime where sparse attention is needed. Therefore, MASA improves accuracy without weakening the practical efficiency of the underlying sparse attention method.

\vspace{-2mm}
\subsection{Ablation Study}
\vspace{-2mm}
MASA is derived from the matrix-product approximation objective in Sec.~\ref{sec:masa}. The remaining implementation choice is the probe matrix $\mX$. The direct choice $\mX=\mV$ gives the attention-output objective, while Sec.~\ref{sec:masa_practical} also introduces norm-controlled probes to reduce the effect of high-norm value outliers. We ablate soft norm balancing in~\eqref{eq:masa_v_norm_alpha} and norm clipping in~\eqref{eq:masa_v_norm_clip}, with full length- and task-level results in Appendix~\ref{app:ablation}.

Together with the main results in Sec.~\ref{sec:main_results}, the probe ablation in Table~\ref{tab:probe_ablation_main} supports two conclusions. First, raw MASA already improves the original sparse baseline in several settings, which supports the matrix-product scoring objective itself. Second, norm control can further improve the score in settings where value-norm outliers affect sparse-unit ranking. Norm clipping is helpful for SeerAttention and FlexPrefill-Llama, while soft norm balancing is useful in some FlexPrefill settings. These trends match the probe design: MASA should preserve value directions while preventing a small number of high-norm value vectors from dominating the selection score.

\begin{wraptable}{R}{0.50\textwidth}
\vspace{-11mm}
\centering
\caption{
Ablation of \masa{} probe choices.
Clip $p$ denotes norm clipping at percentile $p$, and Norm $\alpha$ denotes soft norm balancing.
The case $p=\text{--}$ and $\alpha=\text{--}$ is raw \masa{} without norm control.
Bold marks the best \masa{} variant for each setting. See details in Sec.~\ref{sec:masa_practical}.
}
\vspace{2pt}
\label{tab:probe_ablation_main}
\small
\setlength{\tabcolsep}{4.5pt}
\renewcommand{\arraystretch}{1.06}
\begin{tabular}{@{}C{0.34cm} L{2.35cm} c c C{0.72cm}@{}}
\toprule
& Setting & Clip $p$ & Norm $\alpha$ & \avgcell{Avg.} \\
\midrule

\multirow{12}{*}{\benchlabel{RULER}}
& MInference        & \dash & \dash & \avgcell{87.13} \\
& MInference        & 90    & 0.6   & \avgcell{\best{87.27}} \\
\cmidrule(lr){2-5}

& SeerAttention     & \dash & \dash & \avgcell{87.88} \\
& SeerAttention     & 90    & \dash & \avgcell{\best{88.09}} \\
& SeerAttention     & 95    & \dash & \avgcell{87.79} \\
\cmidrule(lr){2-5}

& FlexPrefill-Llama & 90    & \dash & \avgcell{89.26} \\
& FlexPrefill-Llama & 95    & \dash & \avgcell{89.22} \\
& FlexPrefill-Llama & 97.5 & \dash & \avgcell{\best{89.94}} \\
\cmidrule(lr){2-5}

& FlexPrefill-Qwen  & \dash & \dash & \avgcell{\best{72.16}} \\
& FlexPrefill-Qwen  & 95    & \dash & \avgcell{70.99} \\

\midrule

\multirow{6}{*}{\benchlabel{InfiniteBench}}
& MInference        & \dash & \dash & \avgcell{39.05} \\
& MInference        & 80    & 0.7   & \avgcell{\best{39.20}} \\
\cmidrule(lr){2-5}

& FlexPrefill-Llama & \dash & 0.5   & \avgcell{\best{47.99}} \\
\cmidrule(lr){2-5}

& FlexPrefill-Qwen  & \dash & \dash & \avgcell{\best{30.42}} \\
& FlexPrefill-Qwen  & \dash & 0.5   & \avgcell{30.31} \\

\bottomrule
\end{tabular}
\vspace{-2mm}
\end{wraptable}

\vspace{-2mm}
\section{Conclusion}
\vspace{-2mm}
This paper identifies a fundamental selection mismatch in sparse attention. Existing methods usually retain large attention entries or high-mass regions, which treats the attention matrix as a bag of scalar values. However, Transformers use the attention matrix through matrix multiplication with value vectors. Therefore, sparse attention should be formulated as matrix approximation, not as entrywise attention compression. Based on this view, we propose MASA, a theory-grounded plug-in correction that ranks sparse units by their closed-form reduction in matrix-product approximation error. MASA does not change the sparse kernels, candidate sparse units, or budget rules of existing methods, making it orthogonal to low-level attention kernels and other LLM acceleration techniques as discussed in Appendix~\ref{app:llm_acceleration}. Across multiple sparse baselines, benchmarks, and LLM backbones, MASA gives consistent gains under matched settings with negligible latency overhead. These results support both the proposed method and the broader matrix-approximation view of sparse attention.

One practical boundary of MASA is that it corrects the selection score within a given sparse attention framework. It does not search for new sparse kernels, new sparse unit families, or new budget schedules. This is also what makes the comparison clean: MASA isolates the effect of the selection objective while keeping the original sparse attention system unchanged. Future work can combine MASA with better kernel designs, adaptive budget policies, and other LLM acceleration methods, but the core message remains independent of these engineering choices: sparse attention should preserve the matrix action of attention, not blindly keep large scalar entries.

\newpage

\section*{AI use statement}
LLMs are used to assist with grammar checking, improving writing fluency, managing computational resources and providing suggestions after the initial draft. 
All observations, ideas, methodologies in this paper are developed entirely by the authors. 
Any content generated with the help of LLMs is created under detailed author instructions and thoroughly verified by the authors before inclusion.

\bibliography{references}

\begin{thebibliography}{44}
\providecommand{\natexlab}[1]{#1}
\providecommand{\url}[1]{\texttt{#1}}
\expandafter\ifx\csname urlstyle\endcsname\relax
  \providecommand{\doi}[1]{doi: #1}\else
  \providecommand{\doi}{doi: \begingroup \urlstyle{rm}\Url}\fi

\bibitem[Ahn et~al.(2024)Ahn, Verma, Lou, Liu, Zhang, and Yin]{ahn2024large}
Janice Ahn, Rishu Verma, Renze Lou, Di~Liu, Rui Zhang, and Wenpeng Yin.
\newblock Large language models for mathematical reasoning: Progresses and challenges.
\newblock In \emph{Proceedings of the 18th Conference of the European Chapter of the Association for Computational Linguistics: Student Research Workshop}, pp.\  225--237, 2024.

\bibitem[Bai et~al.(2024{\natexlab{a}})Bai, Chai, Ling, Wang, Lu, Zhang, Shi, Yu, Zhu, Zhang, et~al.]{bai2024beyond}
Guangji Bai, Zheng Chai, Chen Ling, Shiyu Wang, Jiaying Lu, Nan Zhang, Tingwei Shi, Ziyang Yu, Mengdan Zhu, Yifei Zhang, et~al.
\newblock Beyond efficiency: A systematic survey of resource-efficient large language models.
\newblock \emph{arXiv preprint arXiv:2401.00625}, 2024{\natexlab{a}}.

\bibitem[Bai et~al.(2024{\natexlab{b}})Bai, Lv, Zhang, Lyu, Tang, Huang, Du, Liu, Zeng, Hou, et~al.]{bai2024longbench}
Yushi Bai, Xin Lv, Jiajie Zhang, Hongchang Lyu, Jiankai Tang, Zhidian Huang, Zhengxiao Du, Xiao Liu, Aohan Zeng, Lei Hou, et~al.
\newblock Longbench: A bilingual, multitask benchmark for long context understanding.
\newblock In \emph{Proceedings of the 62nd annual meeting of the association for computational linguistics (volume 1: Long papers)}, pp.\  3119--3137, 2024{\natexlab{b}}.

\bibitem[Beltagy et~al.(2020)Beltagy, Peters, and Cohan]{beltagy2020longformer}
Iz~Beltagy, Matthew~E Peters, and Arman Cohan.
\newblock Longformer: The long-document transformer.
\newblock \emph{arXiv preprint arXiv:2004.05150}, 2020.

\bibitem[Bondarenko et~al.(2023)Bondarenko, Nagel, and Blankevoort]{bondarenko2023quantizable}
Yelysei Bondarenko, Markus Nagel, and Tijmen Blankevoort.
\newblock Quantizable transformers: Removing outliers by helping attention heads do nothing.
\newblock \emph{Advances in Neural Information Processing Systems}, 36:\penalty0 75067--75096, 2023.

\bibitem[Borgeaud et~al.(2022)Borgeaud, Mensch, Hoffmann, Cai, Rutherford, Millican, Van Den~Driessche, Lespiau, Damoc, Clark, et~al.]{borgeaud2022improving}
Sebastian Borgeaud, Arthur Mensch, Jordan Hoffmann, Trevor Cai, Eliza Rutherford, Katie Millican, George~Bm Van Den~Driessche, Jean-Baptiste Lespiau, Bogdan Damoc, Aidan Clark, et~al.
\newblock Improving language models by retrieving from trillions of tokens.
\newblock In \emph{International conference on machine learning}, pp.\  2206--2240. PMLR, 2022.

\bibitem[Child et~al.(2019)Child, Gray, Radford, and Sutskever]{child2019generating}
Rewon Child, Scott Gray, Alec Radford, and Ilya Sutskever.
\newblock Generating long sequences with sparse transformers.
\newblock \emph{arXiv preprint arXiv:1904.10509}, 2019.

\bibitem[Dai et~al.(2019)Dai, Yang, Yang, Carbonell, Le, and Salakhutdinov]{dai2019transformer}
Zihang Dai, Zhilin Yang, Yiming Yang, Jaime~G Carbonell, Quoc Le, and Ruslan Salakhutdinov.
\newblock Transformer-xl: Attentive language models beyond a fixed-length context.
\newblock In \emph{Proceedings of the 57th annual meeting of the association for computational linguistics}, pp.\  2978--2988, 2019.

\bibitem[Dao et~al.(2022)Dao, Fu, Ermon, Rudra, and R{\'e}]{dao2022flashattention}
Tri Dao, Dan Fu, Stefano Ermon, Atri Rudra, and Christopher R{\'e}.
\newblock Flashattention: Fast and memory-efficient exact attention with io-awareness.
\newblock \emph{Advances in neural information processing systems}, 35:\penalty0 16344--16359, 2022.

\bibitem[Fu et~al.(2025)Fu, Huang, Ning, Zhang, Chen, Wu, Wang, Huang, Li, Yan, et~al.]{fu2025mixture}
Tianyu Fu, Haofeng Huang, Xuefei Ning, Genghan Zhang, Boju Chen, Tianqi Wu, Hongyi Wang, Zixiao Huang, Shiyao Li, Shengen Yan, et~al.
\newblock Mixture of attention spans: Optimizing llm inference efficiency with heterogeneous sliding-window lengths.
\newblock In \emph{Second Conference on Language Modeling}, 2025.

\bibitem[Gao et~al.(2025)Gao, Zeng, Du, Cao, Zhou, Qi, Lai, So, Cao, Yang, et~al.]{gao2024seerattention}
Yizhao Gao, Zhichen Zeng, Dayou Du, Shijie Cao, Peiyuan Zhou, Jiaxing Qi, Junjie Lai, Hayden Kwok-Hay So, Ting Cao, Fan Yang, et~al.
\newblock Seerattention: Learning intrinsic sparse attention in your llms.
\newblock \emph{Advances in Neural Information Processing Systems}, 2025.

\bibitem[Goel et~al.(2025)Goel, Park, Gagrani, Jones, Morse, Langston, Lee, and Lott]{goel2025caote}
Raghavv Goel, Junyoung Park, Mukul Gagrani, Dalton Jones, Matthew Morse, Harper Langston, Mingu Lee, and Chris Lott.
\newblock Caote: Kv cache selection for llms via attention output error-based token eviction.
\newblock \emph{arXiv preprint arXiv:2504.14051}, 2025.

\bibitem[Gu \& Dao(2023)Gu and Dao]{gu2023mamba}
Albert Gu and Tri Dao.
\newblock Mamba: Linear-time sequence modeling with selective state spaces.
\newblock \emph{arXiv preprint arXiv:2312.00752}, 2023.

\bibitem[Guo et~al.(2024)Guo, Kamigaito, and Watanabe]{guo2024attention}
Zhiyu Guo, Hidetaka Kamigaito, and Taro Watanabe.
\newblock Attention score is not all you need for token importance indicator in kv cache reduction: Value also matters.
\newblock In \emph{Proceedings of the 2024 Conference on Empirical Methods in Natural Language Processing}, pp.\  21158--21166, 2024.

\bibitem[Gupta \& Berant(2021)Gupta and Berant]{gupta2021value}
Ankit Gupta and Jonathan Berant.
\newblock Value-aware approximate attention.
\newblock In \emph{Proceedings of the 2021 Conference on Empirical Methods in Natural Language Processing}, pp.\  9567--9574, 2021.

\bibitem[Hong et~al.(2024)Hong, Dai, Xu, Mao, Li, Liu, Chen, Dong, and Wang]{hong2024flashdecoding++}
Ke~Hong, Guohao Dai, Jiaming Xu, Qiuli Mao, Xiuhong Li, Jun Liu, Kangdi Chen, Yuhan Dong, and Yu~Wang.
\newblock Flashdecoding++: Faster large language model inference with asynchronization, flat gemm optimization, and heuristics.
\newblock \emph{Proceedings of Machine Learning and Systems}, 6:\penalty0 148--161, 2024.

\bibitem[Hsieh et~al.(2024)Hsieh, Sun, Kriman, Acharya, Rekesh, Jia, Zhang, and Ginsburg]{hsieh2024ruler}
Cheng-Ping Hsieh, Simeng Sun, Samuel Kriman, Shantanu Acharya, Dima Rekesh, Fei Jia, Yang Zhang, and Boris Ginsburg.
\newblock Ruler: What's the real context size of your long-context language models?
\newblock \emph{arXiv preprint arXiv:2404.06654}, 2024.

\bibitem[Jiang et~al.(2023{\natexlab{a}})Jiang, Sablayrolles, Mensch, Bamford, Chaplot, de~las Casas, Bressand, Lengyel, Lample, Saulnier, Lavaud, Lachaux, Stock, Scao, Lavril, Wang, Lacroix, and Sayed]{jiang2023mistral7b}
Albert~Q. Jiang, Alexandre Sablayrolles, Arthur Mensch, Chris Bamford, Devendra~Singh Chaplot, Diego de~las Casas, Florian Bressand, Gianna Lengyel, Guillaume Lample, Lucile Saulnier, Lélio~Renard Lavaud, Marie-Anne Lachaux, Pierre Stock, Teven~Le Scao, Thibaut Lavril, Thomas Wang, Timothée Lacroix, and William~El Sayed.
\newblock Mistral 7b, 2023{\natexlab{a}}.
\newblock URL \url{https://arxiv.org/abs/2310.06825}.

\bibitem[Jiang et~al.(2023{\natexlab{b}})Jiang, Wu, Lin, Yang, and Qiu]{jiang2023llmlingua}
Huiqiang Jiang, Qianhui Wu, Chin-Yew Lin, Yuqing Yang, and Lili Qiu.
\newblock Llmlingua: Compressing prompts for accelerated inference of large language models.
\newblock In \emph{Proceedings of the 2023 conference on empirical methods in natural language processing}, pp.\  13358--13376, 2023{\natexlab{b}}.

\bibitem[Jiang et~al.(2024{\natexlab{a}})Jiang, Li, Zhang, Wu, Luo, Ahn, Han, Abdi, Li, Lin, et~al.]{jiang2024minference}
Huiqiang Jiang, Yucheng Li, Chengruidong Zhang, Qianhui Wu, Xufang Luo, Surin Ahn, Zhenhua Han, Amir~H Abdi, Dongsheng Li, Chin-Yew Lin, et~al.
\newblock Minference 1.0: Accelerating pre-filling for long-context llms via dynamic sparse attention.
\newblock \emph{Advances in Neural Information Processing Systems}, 37:\penalty0 52481--52515, 2024{\natexlab{a}}.

\bibitem[Jiang et~al.(2024{\natexlab{b}})Jiang, Wu, Luo, Li, Lin, Yang, and Qiu]{jiang2024longllmlingua}
Huiqiang Jiang, Qianhui Wu, Xufang Luo, Dongsheng Li, Chin-Yew Lin, Yuqing Yang, and Lili Qiu.
\newblock Longllmlingua: Accelerating and enhancing llms in long context scenarios via prompt compression.
\newblock In \emph{Proceedings of the 62nd Annual Meeting of the Association for Computational Linguistics (Volume 1: Long Papers)}, pp.\  1658--1677, 2024{\natexlab{b}}.

\bibitem[Kobayashi et~al.(2020)Kobayashi, Kuribayashi, Yokoi, and Inui]{kobayashi2020attention}
Goro Kobayashi, Tatsuki Kuribayashi, Sho Yokoi, and Kentaro Inui.
\newblock Attention is not only a weight: Analyzing transformers with vector norms.
\newblock In \emph{Proceedings of the 2020 Conference on Empirical Methods in Natural Language Processing (EMNLP)}, pp.\  7057--7075, 2020.

\bibitem[Kwon et~al.(2023)Kwon, Li, Zhuang, Sheng, Zheng, Yu, Gonzalez, Zhang, and Stoica]{kwon2023efficient}
Woosuk Kwon, Zhuohan Li, Siyuan Zhuang, Ying Sheng, Lianmin Zheng, Cody~Hao Yu, Joseph Gonzalez, Hao Zhang, and Ion Stoica.
\newblock Efficient memory management for large language model serving with pagedattention.
\newblock In \emph{Proceedings of the 29th symposium on operating systems principles}, pp.\  611--626, 2023.

\bibitem[Lai et~al.(2025)Lai, Lu, Luo, Ma, and Zhou]{laiflexprefill}
Xunhao Lai, Jianqiao Lu, Yao Luo, Yiyuan Ma, and Xun Zhou.
\newblock Flexprefill: A context-aware sparse attention mechanism for efficient long-sequence inference.
\newblock In \emph{The Thirteenth International Conference on Learning Representations}, 2025.

\bibitem[Liu et~al.(2024{\natexlab{a}})Liu, Zaharia, and Abbeel]{liu2024ringattention}
Hao Liu, Matei Zaharia, and Pieter Abbeel.
\newblock Ringattention with blockwise transformers for near-infinite context.
\newblock In \emph{International Conference on Learning Representations}, volume 2024, pp.\  3992--4008, 2024{\natexlab{a}}.

\bibitem[Liu et~al.(2025)Liu, Yan, Zaharia, and Abbeel]{liu2025world}
Hao Liu, Wilson Yan, Matei Zaharia, and Pieter Abbeel.
\newblock World model on million-length video and language with blockwise ringattention.
\newblock In \emph{International Conference on Learning Representations}, volume 2025, pp.\  45953--45977, 2025.

\bibitem[Liu et~al.(2024{\natexlab{b}})Liu, Yuan, Jin, Zhong, Xu, Braverman, Chen, and Hu]{liu2024kivi}
Zirui Liu, Jiayi Yuan, Hongye Jin, Shaochen Zhong, Zhaozhuo Xu, Vladimir Braverman, Beidi Chen, and Xia Hu.
\newblock Kivi: A tuning-free asymmetric 2bit quantization for kv cache.
\newblock \emph{arXiv preprint arXiv:2402.02750}, 2024{\natexlab{b}}.

\bibitem[Radford et~al.(2018)Radford, Narasimhan, Salimans, Sutskever, et~al.]{radford2018improving}
Alec Radford, Karthik Narasimhan, Tim Salimans, Ilya Sutskever, et~al.
\newblock Improving language understanding by generative pre-training, 2018.

\bibitem[Sengupta et~al.(2026)Sengupta, Chaudhary, and Chakraborty]{sengupta2026value}
Ayan Sengupta, Siddhant Chaudhary, and Tanmoy Chakraborty.
\newblock Value-guided kv compression for llms via approximated cur decomposition.
\newblock \emph{Advances in Neural Information Processing Systems}, 38:\penalty0 79461--79492, 2026.

\bibitem[Sun et~al.(2023)Sun, Dong, Huang, Ma, Xia, Xue, Wang, and Wei]{sun2023retentive}
Yutao Sun, Li~Dong, Shaohan Huang, Shuming Ma, Yuqing Xia, Jilong Xue, Jianyong Wang, and Furu Wei.
\newblock Retentive network: A successor to transformer for large language models.
\newblock \emph{arXiv preprint arXiv:2307.08621}, 2023.

\bibitem[Tang et~al.(2024)Tang, Zhao, Zhu, Xiao, Kasikci, and Han]{tang2024quest}
Jiaming Tang, Yilong Zhao, Kan Zhu, Guangxuan Xiao, Baris Kasikci, and Song Han.
\newblock Quest: Query-aware sparsity for efficient long-context llm inference.
\newblock In \emph{International Conference on Machine Learning}, pp.\  47901--47911. PMLR, 2024.

\bibitem[Team et~al.(2023)Team, Anil, Borgeaud, Alayrac, Yu, Soricut, Schalkwyk, Dai, Hauth, Millican, et~al.]{team2023gemini}
Gemini Team, Rohan Anil, Sebastian Borgeaud, Jean-Baptiste Alayrac, Jiahui Yu, Radu Soricut, Johan Schalkwyk, Andrew~M Dai, Anja Hauth, Katie Millican, et~al.
\newblock Gemini: a family of highly capable multimodal models.
\newblock \emph{arXiv preprint arXiv:2312.11805}, 2023.

\bibitem[Tillet et~al.(2019)Tillet, Kung, and Cox]{tillet2019triton}
Philippe Tillet, Hsiang-Tsung Kung, and David Cox.
\newblock Triton: an intermediate language and compiler for tiled neural network computations.
\newblock In \emph{Proceedings of the 3rd ACM SIGPLAN International Workshop on Machine Learning and Programming Languages}, pp.\  10--19, 2019.

\bibitem[Vaswani et~al.(2017)Vaswani, Shazeer, Parmar, Uszkoreit, Jones, Gomez, Kaiser, and Polosukhin]{vaswani2017attention}
Ashish Vaswani, Noam Shazeer, Niki Parmar, Jakob Uszkoreit, Llion Jones, Aidan~N Gomez, {\L}ukasz Kaiser, and Illia Polosukhin.
\newblock Attention is all you need.
\newblock \emph{Advances in neural information processing systems}, 30, 2017.

\bibitem[Xiao et~al.(2024{\natexlab{a}})Xiao, Zhang, Han, Xiao, Lin, Zhang, Liu, and Sun]{xiao2024infllm}
Chaojun Xiao, Pengle Zhang, Xu~Han, Guangxuan Xiao, Yankai Lin, Zhengyan Zhang, Zhiyuan Liu, and Maosong Sun.
\newblock Infllm: Training-free long-context extrapolation for llms with an efficient context memory.
\newblock \emph{Advances in neural information processing systems}, 37:\penalty0 119638--119661, 2024{\natexlab{a}}.

\bibitem[Xiao et~al.(2024{\natexlab{b}})Xiao, Tian, Chen, Han, and Lewis]{xiao2024efficient}
Guangxuan Xiao, Yuandong Tian, Beidi Chen, Song Han, and Mike Lewis.
\newblock Efficient streaming language models with attention sinks.
\newblock In \emph{International Conference on Learning Representations}, volume 2024, pp.\  21875--21895, 2024{\natexlab{b}}.

\bibitem[Xiao et~al.(2025)Xiao, Tang, Zuo, Guo, Yang, Tang, Fu, and Han]{xiao2025duoattention}
Guangxuan Xiao, Jiaming Tang, Jingwei Zuo, Junxian Guo, Shang Yang, Haotian Tang, Yao Fu, and Song Han.
\newblock Duoattention: Efficient long-context llm inference with retrieval and streaming heads.
\newblock In \emph{International Conference on Learning Representations}, volume 2025, pp.\  37228--37253, 2025.

\bibitem[Xu et~al.(2025)Xu, Xiao, Huang, Guo, and Han]{xuxattention}
Ruyi Xu, Guangxuan Xiao, Haofeng Huang, Junxian Guo, and Song Han.
\newblock Xattention: Block sparse attention with antidiagonal scoring.
\newblock In \emph{Forty-second International Conference on Machine Learning}, 2025.

\bibitem[Zaheer et~al.(2020)Zaheer, Guruganesh, Dubey, Ainslie, Alberti, Ontanon, Pham, Ravula, Wang, Yang, et~al.]{zaheer2020big}
Manzil Zaheer, Guru Guruganesh, Kumar~Avinava Dubey, Joshua Ainslie, Chris Alberti, Santiago Ontanon, Philip Pham, Anirudh Ravula, Qifan Wang, Li~Yang, et~al.
\newblock Big bird: Transformers for longer sequences.
\newblock \emph{Advances in neural information processing systems}, 33:\penalty0 17283--17297, 2020.

\bibitem[Zhang et~al.(2025{\natexlab{a}})Zhang, Xiang, Huang, Xi, Zhu, Chen, et~al.]{zhangspargeattention}
Jintao Zhang, Chendong Xiang, Haofeng Huang, Haocheng Xi, Jun Zhu, Jianfei Chen, et~al.
\newblock Spargeattention: Accurate and training-free sparse attention accelerating any model inference.
\newblock In \emph{Forty-second International Conference on Machine Learning}, 2025{\natexlab{a}}.

\bibitem[Zhang et~al.(2024)Zhang, Chen, Hu, Xu, Chen, Hao, Han, Thai, Wang, Liu, et~al.]{zhang2024bench}
Xinrong Zhang, Yingfa Chen, Shengding Hu, Zihang Xu, Junhao Chen, Moo Hao, Xu~Han, Zhen Thai, Shuo Wang, Zhiyuan Liu, et~al.
\newblock \ensuremath{\infty} bench: Extending long context evaluation beyond 100k tokens.
\newblock In \emph{Proceedings of the 62nd Annual Meeting of the Association for Computational Linguistics (Volume 1: Long Papers)}, pp.\  15262--15277, 2024.

\bibitem[Zhang et~al.(2025{\natexlab{b}})Zhang, Wang, Helwig, Luo, Fu, Xie, Liu, Lin, Xu, Yan, Adams, Weiler, Li, Fu, Wang, Yu, Xie, Fu, Strasser, Xu, Liu, Du, Saxton, Ling, Lawrence, St{\"a}rk, Gui, Edwards, Gao, Ladera, Wu, Hofgard, Tehrani, Wang, Daigavane, Bohde, Kurtin, Huang, Phung, Xu, Joshi, Mathis, Azizzadenesheli, Fang, Aspuru-Guzik, Bekkers, Bronstein, Zitnik, Anandkumar, Ermon, Li{\`o}, Yu, G{\"u}nnemann, Leskovec, Ji, Sun, Barzilay, Jaakkola, Coley, Qian, Qian, Smidt, and Ji]{zhang2025artificial}
Xuan Zhang, Limei Wang, Jacob Helwig, Youzhi Luo, Cong Fu, Yaochen Xie, Meng Liu, Yuchao Lin, Zhao Xu, Keqiang Yan, Keir Adams, Maurice Weiler, Xiner Li, Tianfan Fu, Yucheng Wang, Haiyang Yu, YuQing Xie, Xiang Fu, Alex Strasser, Shenglong Xu, Yi~Liu, Yuanqi Du, Alexandra Saxton, Hongyi Ling, Hannah Lawrence, Hannes St{\"a}rk, Shurui Gui, Carl Edwards, Nicholas Gao, Adriana Ladera, Tailin Wu, Elyssa~F. Hofgard, Aria~Mansouri Tehrani, Rui Wang, Ameya Daigavane, Montgomery Bohde, Jerry Kurtin, Qian Huang, Tuong Phung, Minkai Xu, Chaitanya~K. Joshi, Simon~V. Mathis, Kamyar Azizzadenesheli, Ada Fang, Al{\'a}n Aspuru-Guzik, Erik Bekkers, Michael Bronstein, Marinka Zitnik, Anima Anandkumar, Stefano Ermon, Pietro Li{\`o}, Rose Yu, Stephan G{\"u}nnemann, Jure Leskovec, Heng Ji, Jimeng Sun, Regina Barzilay, Tommi Jaakkola, Connor~W. Coley, Xiaoning Qian, Xiaofeng Qian, Tess Smidt, and Shuiwang Ji.
\newblock Artificial intelligence for science in quantum, atomistic, and continuum systems.
\newblock \emph{Foundations and Trends{\textregistered} in Machine Learning}, 18\penalty0 (4):\penalty0 385--912, 2025{\natexlab{b}}.
\newblock ISSN 1935-8237.
\newblock \doi{10.1561/2200000115}.
\newblock URL \url{http://dx.doi.org/10.1561/2200000115}.

\bibitem[Zhang et~al.(2023)Zhang, Sheng, Zhou, Chen, Zheng, Cai, Song, Tian, R{\'e}, Barrett, et~al.]{zhang2023h2o}
Zhenyu Zhang, Ying Sheng, Tianyi Zhou, Tianlong Chen, Lianmin Zheng, Ruisi Cai, Zhao Song, Yuandong Tian, Christopher R{\'e}, Clark Barrett, et~al.
\newblock H2o: Heavy-hitter oracle for efficient generative inference of large language models.
\newblock \emph{Advances in Neural Information Processing Systems}, 36:\penalty0 34661--34710, 2023.

\bibitem[Zhu et~al.(2025)Zhu, Tang, Xu, Gu, Zeng, Kadekodi, Zhao, Li, Krishnamurthy, and Kasikci]{zhu2025tactic}
Kan Zhu, Tian Tang, Qinyu Xu, Yile Gu, Zhichen Zeng, Rohan Kadekodi, Liangyu Zhao, Ang Li, Arvind Krishnamurthy, and Baris Kasikci.
\newblock Tactic: Adaptive sparse attention with clustering and distribution fitting for long-context llms.
\newblock \emph{arXiv preprint arXiv:2502.12216}, 2025.

\end{thebibliography}
\bibliographystyle{masa}

\newpage

\appendix

\section{LLM Acceleration}
\label{app:llm_acceleration}

\textbf{LLM inference acceleration.}
The quadratic cost of self-attention with respect to sequence length has motivated many acceleration methods for long-context LLM inference. One line improves hardware efficiency through IO-aware or distributed attention kernels, including FlashAttention~\citep{dao2022flashattention}, FlashDecoding++~\citep{hong2024flashdecoding++}, and RingAttention~\citep{liu2024ringattention}. Another reduces the effective context length through retrieval, prompt compression, or context-memory mechanisms~\citep{borgeaud2022improving,jiang2023llmlingua,jiang2024longllmlingua,xiao2024infllm}. For decoding, methods address KV-cache efficiency through paged memory management, quantization, selective cache use, and eviction~\citep{kwon2023efficient,liu2024kivi,xiao2025duoattention,zhang2023h2o}. Other approaches change the model architecture through recurrence, state-space models, linear attention, or hybrid designs~\citep{dai2019transformer,sun2023retentive,gu2023mamba}. These approaches reduce inference cost through computation, memory, context, or architectural changes.

\textbf{Oracle analysis and KV-cache reduction.}
\citet{gupta2021value} analyze per-query oracle approximations expressed as sparse convex combinations of value vectors, setting aside computational and memory constraints. VATP~\citep{guo2024attention}, CAOTE~\citep{goel2025caote}, and CurDKV~\citep{sengupta2026value} instead address KV-cache reduction, selecting retained states through norm-weighted token importance, eviction-induced output error, and CUR-based scores, respectively. Their selection objects are approximation coefficients or cached KV states. MASA instead ranks framework-defined masks over query--key interactions for structured sparse attention computation.

\textbf{Sparse attention.} Sparse attention reduces computation by retaining only a subset of query--key interactions~\citep{child2019generating,beltagy2020longformer,zaheer2020big,xiao2024efficient}. Methods differ in how they construct this subset: fixed local, global, or sink patterns; offline assignment of structured patterns; learned prediction of input-dependent regions; and adaptive selection budgets~\citep{jiang2024minference,gao2024seerattention,laiflexprefill,zhu2025tactic}. The representative sparse-prefill frameworks reviewed in Sec.~\ref{sec:baselines} rank candidate units using attention-weight summaries or learned attention-pattern targets. These criteria estimate where attention is concentrated, rather than directly evaluating the matrix-product error reduction from retaining a candidate unit. Thus, constructing a sparse pattern family and choosing how to rank its units are distinct design questions.

\textbf{Position and novelty of MASA.}
MASA is orthogonal to sparse-kernel optimization and
KV-cache storage design: \textbf{it changes the selection criterion, not the execution kernel or cache representation.} Within sparse attention, MASA also retains the candidate pattern family and budget rule. Pattern construction defines the admissible sparse units, the budget constrains their selection, and the kernel executes the selected interactions. MASA addresses a different question: \emph{which units should be retained to preserve the attention matrix product?}

MASA answers this question through \textbf{matrix-approximation ranking of structured sparse units}. The closed-form score in~\eqref{eq:masa_unit_score} measures the reduction in retained matrix-product error from selecting each candidate individually. It accounts for alignment with the target product $\mA\mX$ and interactions among entries within the unit, yielding a common criterion for vertical, slash, and block patterns. The implementation evaluates this criterion at the framework's existing profile resolution through profile-matched probes and memory-safe matrix products (Sec.~\ref{sec:masa_practical}). Raw, pooled, and norm-controlled probes use the same scoring formulation without changing the value vectors consumed by the sparse kernel.

\textbf{This distinction is the main novelty of MASA:
it connects a structured matrix-approximation objective
to a common ranking mechanism within existing sparse
attention systems.}
Rather than introducing another sparse pattern, compression
policy, or execution kernel, MASA changes what makes a
candidate unit important: its contribution to matrix-product
preservation, not merely its attention mass.
It therefore complements advances in sparse execution
and memory management with a theory-grounded correction
to the selection principle.

\section{Frobenius Norm and Matrix-Product Error}
\label{app:frobenius_norm}

For a matrix $\mZ\in\mathbb{R}^{m\times n}$, the squared Frobenius norm is
\begin{equation}
\label{eq:app_frobenius_definition}
\|\mZ\|_F^2
=
\sum_{i=1}^{m}\sum_{j=1}^{n}\mZ_{ij}^{2}
=
\operatorname{tr}(\mZ\mZ^{\top}) .
\end{equation}
It measures the total squared error over all matrix entries. In sparse attention,
the attention output has one row per query position, so
\begin{equation}
\left\|
\mA\mV
-
(\mM_{\mathcal{S}}\odot\mA)\mV
\right\|_F^2
\end{equation}
measures the total squared output error over all query positions and value
dimensions.

For a general probe matrix $\mX$, the MASA loss is
\begin{equation}
\label{eq:app_action_frobenius}
\mathcal{L}_{\mathrm{MASA}}(\mathcal{S};\mX)
=
\left\|
\left(\mA-\mM_{\mathcal{S}}\odot\mA\right)\mX
\right\|_F^2 .
\end{equation}
Using~\eqref{eq:app_frobenius_definition}, this expands to
\begin{equation}
\label{eq:app_action_trace}
\mathcal{L}_{\mathrm{MASA}}(\mathcal{S};\mX)
=
\operatorname{tr}
\left(
\left(\mA-\mM_{\mathcal{S}}\odot\mA\right)
\mX\mX^{\top}
\left(\mA-\mM_{\mathcal{S}}\odot\mA\right)^{\top}
\right).
\end{equation}
This is the form used in Sec.~\ref{sec:masa}. It shows that MASA does not measure
the entrywise error of $\mA$ alone. It measures the error after the attention matrix
acts on $\mX$.

\section{Metric View of MASA}
\label{app:masa_metric}

The probe matrix $\mX$ specifies which matrix product should be preserved. The
direct choice $\mX=\mV$ gives the standard attention-output error:
\begin{equation}
\label{eq:app_value_probe}
\mathcal{L}_{\mathrm{MASA}}(\mathcal{S};\mV)
=
\left\|
\mA\mV
-
(\mM_{\mathcal{S}}\odot\mA)\mV
\right\|_F^2 ,
\end{equation}
which is exactly the objective in~\eqref{eq:objective_budget}.

More general output metrics can also be written in the same form. Let
$\mH\succeq 0$ be a positive semidefinite metric over the value dimension, and
define
\begin{equation}
\label{eq:app_weighted_norm}
\|\mZ\|_{\mH}^{2}
=
\operatorname{tr}(\mZ\mH\mZ^{\top}) .
\end{equation}
Since $\mH\succeq 0$, there exists a matrix $\mR$ such that
\begin{equation}
\label{eq:app_metric_factorization}
\mH=\mR\mR^{\top}.
\end{equation}
Then the weighted attention-output error satisfies
\begin{align}
\label{eq:app_weighted_metric}
\left\|
\left(\mA-\mM_{\mathcal{S}}\odot\mA\right)\mV
\right\|_{\mH}^{2}
&=
\operatorname{tr}
\left(
\left(\mA-\mM_{\mathcal{S}}\odot\mA\right)
\mV
\mH
\mV^{\top}
\left(\mA-\mM_{\mathcal{S}}\odot\mA\right)^{\top}
\right) \\
&=
\operatorname{tr}
\left(
\left(\mA-\mM_{\mathcal{S}}\odot\mA\right)
\mV
\mR\mR^{\top}
\mV^{\top}
\left(\mA-\mM_{\mathcal{S}}\odot\mA\right)^{\top}
\right) \\
&=
\left\|
\left(\mA-\mM_{\mathcal{S}}\odot\mA\right)
\mV\mR
\right\|_F^2 .
\end{align}
Thus, choosing
\begin{equation}
\label{eq:app_weighted_probe}
\mX=\mV\mR
\end{equation}
turns the weighted output error into the MASA objective:
\begin{equation}
\label{eq:app_weighted_masa}
\left\|
\left(\mA-\mM_{\mathcal{S}}\odot\mA\right)\mV
\right\|_{\mH}^{2}
=
\mathcal{L}_{\mathrm{MASA}}(\mathcal{S};\mV\mR).
\end{equation}
This covers raw value probes, projected value probes, and weighted value-space
metrics under the same matrix-product view.

The Frobenius action error also controls the spectral action error. For any matrix
$\mZ$,
\begin{equation}
\label{eq:app_fro_controls_spectral}
\|\mZ\|_{\mathrm{op}}
\le
\|\mZ\|_F .
\end{equation}
Applying this inequality to the MASA residual gives
\begin{equation}
\label{eq:app_action_spectral_bound}
\left\|
\left(\mA-\mM_{\mathcal{S}}\odot\mA\right)\mX
\right\|_{\mathrm{op}}
\le
\left\|
\left(\mA-\mM_{\mathcal{S}}\odot\mA\right)\mX
\right\|_F .
\end{equation}
More generally, for Schatten norms,
\begin{equation}
\label{eq:app_schatten_bound}
\|\mZ\|_q
\le
\|\mZ\|_p,
\qquad
1\le p\le q\le\infty .
\end{equation}
Therefore, reducing the Frobenius matrix-product error also reduces an upper bound
on stronger matrix-action errors such as the spectral norm. MASA uses the
Frobenius objective because it gives the closed-form unit score in
Sec.~\ref{sec:masa}, while still measuring a matrix-product error rather than an
entrywise attention-map error.

\section{Full Tables}
\label{app:full_tables}

The main text reports compact result tables for readability. Here we provide the full comparison layout for each benchmark. Rows marked with $\dagger$ are copied from the corresponding baseline papers and are included only for reference, because they may use different models, sparse budgets, kernels, or evaluation settings. The controlled evidence for MASA comes from the matched rows in each block: the original sparse baseline and the same baseline augmented with MASA. In these matched rows, the model, sparse budget, sparse kernel, sparse unit family, and data processing are kept unchanged; MASA only replaces the sparse-unit ranking score.

Table~\ref{tab:ruler_full} gives the full RULER comparison across sequence lengths from 4K to 128K. Table~\ref{tab:infinitebench_full} gives the full InfiniteBench comparison across summarization, question answering, code, math, and retrieval tasks. Table~\ref{tab:longbench_full} gives the full LongBench comparison across length buckets. These full tables separate reported prior results from the matched MASA comparison, so the conclusion is not based on mixing results from different settings.

\newcommand{\fullfwlabel}[1]{%
  \rotatebox[origin=c]{90}{%
    \fontsize{7.4}{7.7}\selectfont\bfseries #1%
  }%
}
\newcommand{\fullbblabel}[1]{%
  \rotatebox[origin=c]{90}{%
    \fontsize{7.2}{7.5}\selectfont\itshape #1%
  }%
}

\newcommand{\fullfirstsep}{%
  \specialrule{0.90pt}{0pt}{1.3pt}%
}
\newcommand{\fullblocksep}{%
  \specialrule{0.90pt}{1.3pt}{1.3pt}%
}

\newcommand{\rulerfullrefline}{%
  \arrayrulecolor{black!28}\cline{3-9}\arrayrulecolor{black}%
}
\newcommand{\rulerfullpairline}{%
  \arrayrulecolor{black!48}\cline{3-9}\arrayrulecolor{black}%
}
\newcommand{\rulerfullmodelsep}{%
  \arrayrulecolor{black!65}\cline{2-10}\arrayrulecolor{black}%
}

\newcommand{\inffullrefline}{%
  \arrayrulecolor{black!28}\cline{3-13}\arrayrulecolor{black}%
}
\newcommand{\inffullpairline}{%
  \arrayrulecolor{black!48}\cline{3-13}\arrayrulecolor{black}%
}
\newcommand{\inffullmodelsep}{%
  \arrayrulecolor{black!65}\cline{2-14}\arrayrulecolor{black}%
}

\newcommand{\longfullrefline}{%
  \arrayrulecolor{black!28}\cline{1-4}\arrayrulecolor{black}%
}
\newcommand{\longfullpairline}{%
  \arrayrulecolor{black!48}\cline{1-4}\arrayrulecolor{black}%
}

\begin{table}[!htbp]
\centering
\caption{
RULER~\citep{hsieh2024ruler} results with additional methods reported in the
original baseline papers. Rows marked with $\dagger$ are included for context
and may use different settings. In each framework block, Full Attention and
reported methods are followed by the matched Original/\masa{} pair.
\textbf{Bold} marks the better Avg. within each pair.
}
\label{tab:ruler_full}
\small
\setlength{\tabcolsep}{3.6pt}
\renewcommand{\arraystretch}{1.10}

\begin{tabular}{
@{}
C{0.46cm}
@{\hspace{0.5pt}}
C{0.42cm}
@{\hspace{1.3pt}}
l
c c c c c c
!{\color{black!45}\vrule width 0.35pt}
>{\columncolor{AvgCol}\centering\arraybackslash}m{0.78cm}
@{}
}
\toprule
& & Variant & 4K & 8K & 16K & 32K & 64K & 128K & Avg. \\
\fullfirstsep

\multirow[c]{7}{*}{\fullfwlabel{MInfer}}
& & Full Attention
& 97.2 & 91.8 & 87.3 & 80.8 & 77.4 & 72.2 & 84.4 \\
\rulerfullrefline

& & StreamingLLM\reported
& 97.2 & 38.1 & 37.5 & 17.2 & 14.2 & 9.4 & 35.0 \\

& & StreamingLLM w/ dilated\reported
& 23.4 & 0.7 & 1.4 & 18.8 & 16.5 & 15.6 & 12.7 \\

& & StreamingLLM w/ strided\reported
& 2.0 & 0.7 & 0.6 & 0.6 & 0.7 & 1.3 & 1.0 \\

& & InfLLM\reported
& 89.4 & 79.8 & 70.1 & 55.6 & 43.0 & 39.5 & 62.9 \\

\rulerfullpairline

& & \basecell{Original}
& \basecell{97.7} & \basecell{91.2} & \basecell{88.5}
& \basecell{85.0} & \basecell{82.3} & \basecell{77.6}
& 87.0 \\

& & \masacell{\masa}
& \masacell{97.93} & \masacell{91.29} & \masacell{88.50}
& \masacell{85.45} & \masacell{81.57} & \masacell{78.88}
& \best{87.27} \\

\fullblocksep

\multirow[c]{5}{*}{\fullfwlabel{SeerAttn}}
& & Full Attention
& 95.53 & 92.37 & 92.01 & 87.63 & 84.39 & 76.26 & 88.01 \\
\rulerfullrefline

& & MInference\reported
& 95.53 & 92.64 & 91.37 & 85.71 & 83.24 & 67.02 & 85.92 \\

& & DuoAttention\reported
& 95.64 & 92.08 & 90.71 & 84.75 & 83.24 & 75.32 & 86.96 \\

\rulerfullpairline

& & \basecell{Original}
& \basecell{95.53} & \basecell{92.71} & \basecell{92.02}
& \basecell{88.49} & \basecell{83.48} & \basecell{73.37}
& 87.60 \\

& & \masacell{\masa}
& \masacell{95.53} & \masacell{92.74} & \masacell{92.84}
& \masacell{89.56} & \masacell{84.01} & \masacell{73.85}
& \best{88.09} \\

\fullblocksep

\multirow[c]{10}{*}{\fullfwlabel{FlexPrefill}}
& \multirow[c]{5}{*}{\fullbblabel{Llama}}
& Full Attention
& 95.67 & 93.75 & 93.03 & 87.26 & 84.37 & 78.13 & 88.70 \\
\rulerfullrefline

& & StreamingLLM\reported
& 95.43 & 93.99 & 74.76 & 48.56 & 26.20 & 30.77 & 61.62 \\

& & MInference\reported
& 95.67 & 93.99 & 93.27 & 86.54 & 84.86 & 58.17 & 85.42 \\

\rulerfullpairline

& & \basecell{Original}
& \basecell{95.43} & \basecell{93.51} & \basecell{94.71}
& \basecell{89.42} & \basecell{82.93} & \basecell{79.09}
& 89.18 \\

& & \masacell{\masa}
& \masacell{95.43} & \masacell{93.99} & \masacell{94.95}
& \masacell{89.66} & \masacell{85.09} & \masacell{80.54}
& \best{89.94} \\

\rulerfullmodelsep

& \multirow[c]{5}{*}{\fullbblabel{Qwen}}
& Full Attention
& 89.90 & 88.70 & 80.77 & 79.33 & 56.49 & 17.79 & 68.83 \\
\rulerfullrefline

& & StreamingLLM\reported
& 90.14 & 88.94 & 57.93 & 43.03 & 25.48 & 12.26 & 52.96 \\

& & MInference\reported
& 89.90 & 88.70 & 79.33 & 78.61 & 49.04 & 10.82 & 66.07 \\

\rulerfullpairline

& & \basecell{Original}
& \basecell{90.39} & \basecell{89.91} & \basecell{83.17}
& \basecell{81.25} & \basecell{59.14} & \basecell{20.67}
& 70.75 \\

& & \masacell{\masa}
& \masacell{90.88} & \masacell{88.94} & \masacell{84.85}
& \masacell{82.64} & \masacell{62.79} & \masacell{22.83}
& \best{72.16} \\

\bottomrule
\end{tabular}
\end{table}

\FloatBarrier

\begin{table*}[!htbp]
\centering
\caption{
InfiniteBench~\citep{zhang2024bench} results with additional methods reported
in the original baseline papers. Rows marked with $\dagger$ are included for
context and may use different settings. In each framework block, Full Attention
and reported methods are followed by the matched Original/\masa{} pair.
\textbf{Bold} marks the better Avg. within each pair.
}
\label{tab:infinitebench_full}
\scriptsize
\setlength{\tabcolsep}{2.2pt}
\renewcommand{\arraystretch}{1.10}

\resizebox{\textwidth}{!}{%
\begin{tabular}{
@{}
C{0.46cm}
@{\hspace{0.5pt}}
C{0.42cm}
@{\hspace{1.3pt}}
l
c c c c c c c c c c
!{\color{black!45}\vrule width 0.35pt}
>{\columncolor{AvgCol}\centering\arraybackslash}m{0.76cm}
@{}
}
\toprule
& & Variant
& En.Sum & En.QA & En.MC & En.Dia & Zh.QA
& \makecell[c]{Code\\Debug}
& \makecell[c]{Math\\Find}
& \makecell[c]{Retr.\\PassKey}
& \makecell[c]{Retr.\\Number}
& \makecell[c]{Retr.\\KV}
& Avg. \\
\fullfirstsep

\multirow[c]{8}{*}{\fullfwlabel{MInfer}}
& & Full Attention
& 20.2 & 12.4 & 67.3 & 6.0 & 12.9
& 22.1 & 26.6 & 100.0 & 100.0 & 14.4 & 38.2 \\
\inffullrefline

& & StreamingLLM\reported
& 21.0 & 8.2 & 40.2 & 10.0 & 10.4
& 25.9 & 30.0 & 86.8 & 5.1 & 0.8 & 23.8 \\

& & StreamingLLM w/ dilated\reported
& 20.1 & 9.4 & 44.5 & 15.5 & 11.2
& 20.5 & 27.5 & 5.0 & 87.5 & 0.5 & 24.2 \\

& & StreamingLLM w/ strided\reported
& 17.3 & 8.2 & 27.5 & 14.5 & 11.2
& 19.5 & 27.5 & 4.0 & 2.1 & 1.0 & 13.3 \\

& & InfLLM\reported
& 24.1 & 7.8 & 45.0 & 6.0 & 11.4
& 19.5 & 32.9 & 100.0 & 100.0 & 1.2 & 34.8 \\

& & MInference w/ static\reported
& 19.9 & 8.6 & 43.2 & 3.5 & 8.9
& 20.6 & 25.1 & 92.4 & 96.3 & 0.2 & 31.9 \\

\inffullpairline

& & \basecell{Original}
& \basecell{20.5} & \basecell{12.9} & \basecell{65.9}
& \basecell{7.5} & \basecell{12.5} & \basecell{22.3}
& \basecell{33.1} & \basecell{100.0} & \basecell{100.0}
& \basecell{12.8} & 38.8 \\

& & \masacell{\masa}
& \masacell{19.73} & \masacell{12.15} & \masacell{66.81}
& \masacell{8.50} & \masacell{12.28} & \masacell{22.84}
& \masacell{33.43} & \masacell{100.00} & \masacell{100.00}
& \masacell{16.20} & \best{39.20} \\

\fullblocksep

\multirow[c]{10}{*}{\fullfwlabel{FlexPrefill}}
& \multirow[c]{5}{*}{\fullbblabel{Llama}}
& Full Attention
& 31.91 & 25.92 & 69.43 & 21.50 & 31.95
& 16.75 & 24.29 & 99.15 & 99.66 & 60.00 & 48.06 \\
\inffullrefline

& & StreamingLLM\reported
& 30.15 & 10.15 & 41.05 & 8.50 & 22.38
& 8.63 & 17.71 & 2.71 & 5.93 & 0.00 & 14.72 \\

& & MInference\reported
& 31.04 & 22.00 & 63.76 & 14.50 & 28.70
& 5.33 & 27.43 & 56.78 & 77.12 & 14.00 & 34.06 \\

\inffullpairline

& & \basecell{Original}
& \basecell{31.82} & \basecell{24.82} & \basecell{69.43}
& \basecell{19.50} & \basecell{35.46} & \basecell{16.75}
& \basecell{31.14} & \basecell{98.64} & \basecell{99.83}
& \basecell{44.00} & 47.14 \\

& & \masacell{\masa}
& \masacell{32.10} & \masacell{26.25} & \masacell{66.81}
& \masacell{19.50} & \masacell{33.66} & \masacell{20.05}
& \masacell{29.14} & \masacell{98.31} & \masacell{99.66}
& \masacell{54.40} & \best{47.99} \\

\inffullmodelsep

& \multirow[c]{5}{*}{\fullbblabel{Qwen}}
& Full Attention
& 4.65 & 5.57 & 34.50 & 9.00 & 11.27
& 24.87 & 24.57 & 95.42 & 75.42 & 0.00 & 29.53 \\
\inffullrefline

& & StreamingLLM\reported
& 18.54 & 6.43 & 39.30 & 12.50 & 10.94
& 24.11 & 28.00 & 30.85 & 61.69 & 0.00 & 23.24 \\

& & MInference\reported
& 7.45 & 3.94 & 14.85 & 4.50 & 10.17
& 13.20 & 30.00 & 83.90 & 58.47 & 0.00 & 22.65 \\

\inffullpairline

& & \basecell{Original}
& \basecell{14.27} & \basecell{6.55} & \basecell{34.06}
& \basecell{13.50} & \basecell{11.51} & \basecell{20.81}
& \basecell{30.86} & \basecell{96.95} & \basecell{72.20}
& \basecell{0.00} & 30.07 \\

& & \masacell{\masa}
& \masacell{15.55} & \masacell{6.38} & \masacell{32.31}
& \masacell{16.50} & \masacell{13.09} & \masacell{21.07}
& \masacell{30.29} & \masacell{94.41} & \masacell{74.58}
& \masacell{0.00} & \best{30.42} \\

\bottomrule
\end{tabular}%
}
\end{table*}

\begin{table}[!htbp]
\centering
\caption{
LongBench~\citep{bai2024longbench} results with additional methods reported in
the original SeerAttention paper. Rows marked with $\dagger$ are included for
context and may use different settings. Full Attention and reported methods are
followed by the matched Original/\masa{} pair. \textbf{Bold} marks the better
Avg. within the pair.
}
\label{tab:longbench_full}
\small
\setlength{\tabcolsep}{4.8pt}
\renewcommand{\arraystretch}{1.10}

\begin{tabular}{
@{}
l
c c c
!{\color{black!45}\vrule width 0.35pt}
>{\columncolor{AvgCol}\centering\arraybackslash}m{0.80cm}
@{}
}
\toprule
Variant & 0--4K & 4--8K & 8K+ & Avg. \\
\fullfirstsep

Full Attention
& 55.32 & 53.98 & 52.90 & 54.07 \\

\longfullrefline

MInference\reported
& 55.23 & 53.78 & 52.18 & 53.73 \\

MoA\reported
& 50.74 & 49.84 & 51.89 & 50.82 \\

DuoAttention\reported
& 53.77 & 52.17 & 51.27 & 52.40 \\

\longfullpairline

\basecell{Original}
& \basecell{55.43} & \basecell{54.49} & \basecell{52.69}
& 54.20 \\

\masacell{\masa}
& \masacell{55.45} & \masacell{54.48} & \masacell{53.16}
& \best{54.37} \\

\bottomrule
\end{tabular}
\end{table}

\FloatBarrier

\section{Attention-Output Approximation Error}
\label{app:attention_output_error}

\textbf{Direct approximation evidence.}
We complement the downstream results in Sec.~\ref{sec:main_results} by directly measuring how closely the evaluated MASA configurations preserve dense attention outputs. Dense causal FlashAttention-2 and the sparse implementations receive identical $\mQ,\mK,\mV$ tensors. We compare their full-sequence outputs before the attention output projection, including the sparse kernels' softmax normalization.

\textbf{Error metric.}
For each evaluated layer, the normalized squared attention-output error is
\begin{equation}
\label{eq:app_executed_output_error}
E
=
\frac{
\|\mO_{\mathrm{sparse}}-\mO_{\mathrm{dense}}\|_F^2
}{
\|\mO_{\mathrm{dense}}\|_F^2
}.
\end{equation}
The Frobenius norm covers the full output matrix. We average errors across the evaluated layers within each prompt and report the aggregate errors. The relative reduction is $100(E_{\mathrm{Original}}-E_{\mathrm{MASA}})/E_{\mathrm{Original}}$; a positive value indicates improved approximation.

\begin{table}[!htbp]
\centering
\caption{
\textbf{Attention-output error decreases across all four evaluated framework--backbone pairs.} Original and MASA denote the baseline and diagnostic configuration. $E$ is the normalized squared output error in~\eqref{eq:app_executed_output_error}. Relative reductions and paired $98.75\%$ confidence intervals are reported in percent. Bold marks the lower error and the positive reduction.
}
\label{tab:attention_output_error_64k}
\small
\setlength{\tabcolsep}{3.5pt}
\renewcommand{\arraystretch}{1.12}

\begin{tabular}{@{}lllccc@{}}
\toprule
Framework
& Backbone
& Variant
& $E \downarrow$
& \shortstack{Reduction\\(\%) $\uparrow$}
& \shortstack{Paired $98.75\%$\\interval (\%)} \\
\midrule

\multirow{2}{*}{MInference}
& \multirow{2}{*}{Llama}
& Original
& $1.095985$
& --
& -- \\
& & MASA
& $\boldsymbol{1.091532}$
& \textbf{+0.41}
& $[+0.14,\,+0.70]$ \\
\midrule

\multirow{2}{*}{SeerAttention}
& \multirow{2}{*}{Llama}
& Original
& $1.843\times10^{-3}$
& --
& -- \\
& & MASA
& $\boldsymbol{1.805\times10^{-3}}$
& \textbf{+2.10}
& $[+0.69,\,+4.44]$ \\
\midrule

\multirow{4}{*}{FlexPrefill}
& \multirow{2}{*}{Llama}
& Original
& $4.9059\times10^{-2}$
& --
& -- \\
& & MASA
& $\boldsymbol{4.7909\times10^{-2}}$
& \textbf{+2.34}
& $[+0.50,\,+4.18]$ \\
\cmidrule(l){2-6}

& \multirow{2}{*}{Qwen}
& Original
& $4.2009\times10^{-2}$
& --
& -- \\
& & MASA
& $\boldsymbol{3.8997\times10^{-2}}$
& \textbf{+7.17}
& $[+2.33,\,+10.82]$ \\
\bottomrule
\end{tabular}
\vspace{+4mm}
\end{table}

\textbf{Results.}
Table~\ref{tab:attention_output_error_64k} shows \textbf{lower attention-output error in every evaluated framework--backbone pair, with all four reported paired confidence intervals above zero}. The relative reductions are $0.41\%$ for MInference, $2.10\%$ for SeerAttention, and $2.34\%$ and $7.17\%$ for FlexPrefill on Llama and Qwen, respectively. These measurements directly assess the quantity that motivates MASA: preservation of the attention output, rather than attention mass alone. The evaluated MASA configurations produce outputs closer to dense attention, providing empirical support for the matrix-approximation view alongside the downstream accuracy results.

\textbf{Evaluation protocol.}
We use 64K RULER~\citep{hsieh2024ruler} inputs with a $10\%$ retention target and 208 prompts per model, comprising 16 prompts from each of the 13 task families. Dense and sparse attention receive identical post-RoPE $\mQ,\mK,\mV$ tensors. Errors from eight prespecified layers are averaged within each prompt, and uncertainty is estimated using paired hierarchical task/prompt bootstrap. The reported intervals describe these selected-setting comparisons rather than an independent held-out evaluation. All comparisons are within framework--backbone pairs.

\section{Full Ablation Tables}
\label{app:ablation}

\textbf{Probe configurations.}
Probe construction determines how value information enters MASA's matrix-product ranking. We compare raw value probes, norm clipping in~\eqref{eq:masa_v_norm_clip}, and soft norm balancing in~\eqref{eq:masa_v_norm_alpha}, while keeping the sparse baseline, budget, kernel, candidate units, and reduction score in~\eqref{eq:masa_unit_score} fixed. The percentile $p$ determines the clipping threshold within the profile region, while $\alpha$ controls the strength of norm balancing.

For each framework, backbone, and benchmark, the main results use the tested probe configuration with the highest aggregate score on that benchmark. The following tables report the evaluated configurations and their full task- and length-level results.

Table~\ref{tab:app_ruler_probe_ablation} reports the RULER ablation across context lengths. Norm control improves the best tested MInference and SeerAttention results, while its effect varies across FlexPrefill configurations.
Table~\ref{tab:app_infinitebench_probe_ablation} reports the InfiniteBench ablation across tasks, comparing raw and
norm-controlled probes. Table~\ref{tab:app_longbench_probe_ablation} reports the LongBench ablation for SeerAttention, where clipping at $p=87.5$ gives the highest average among the tested configurations. Together, these ablations examine probe construction within MASA's matrix-approximation framework, comparing different treatments of value norms under the same closed-form ranking rule and sparse-attention settings.

\begin{table*}[h]
\centering
\caption{
Full RULER probe ablation.
Clip $p$ denotes norm clipping at percentile $p$, and Norm $\alpha$ denotes soft norm balancing.
The case $p=\text{--}$ and $\alpha=\text{--}$ is raw MASA without norm control.
\textbf{Bold} Avg. marks the best MASA variant within each setting.
}
\label{tab:app_ruler_probe_ablation}
\scriptsize
\setlength{\tabcolsep}{3.2pt}
\renewcommand{\arraystretch}{1.06}
\resizebox{0.9\textwidth}{!}{
\begin{tabular}{llcccccccc}
\toprule
Setting & Clip $p$ & Norm $\alpha$ & 4K & 8K & 16K & 32K & 64K & 128K & \avgcell{Avg.} \\
\midrule

MInference & \dash & \dash
& 97.66 & 91.46 & 88.57 & 85.21 & 81.76 & 78.11 & \avgcell{87.13} \\
MInference & 90 & 0.6
& 97.93 & 91.29 & 88.50
& 85.45 & 81.57 & 78.88
& \avgcell{\best{87.27}} \\
\midrule

SeerAttention & \dash & \dash
& 95.53 & 92.50 & 92.98 & 89.58 & 84.20 & 72.50 & \avgcell{87.88} \\
SeerAttention & 85 & \dash
& 95.63 & 92.40 & 91.98 & 89.55 & 83.64 & 73.34 & \avgcell{87.76} \\
SeerAttention & 90 & \dash
& 95.53 & 92.74 & 92.84 & 89.56 & 84.01 & 73.85 & \avgcell{\best{88.09}} \\
SeerAttention & 95 & \dash
& 95.32 & 92.84 & 92.70 & 89.59 & 83.52 & 72.72 & \avgcell{87.79} \\

\midrule

FlexPrefill-Llama & 90 & \dash
& 95.67 & 94.47 & 93.99 & 88.70 & 84.13 & 78.61 & \avgcell{89.26} \\
FlexPrefill-Llama & 95 & \dash
& 95.67 & 94.23 & 93.99 & 88.94 & 83.89 & 78.61 & \avgcell{89.22} \\
FlexPrefill-Llama & 97.5 & \dash
& 95.43 & 93.99 & 94.95 & 89.66 & 85.09 & 80.54 & \avgcell{\best{89.94}} \\

\midrule

FlexPrefill-Qwen & \dash & \dash
& 90.88 & 88.94 & 84.85 & 82.64 & 62.79 & 22.83 & \avgcell{\best{72.16}} \\
FlexPrefill-Qwen & 95 & \dash
& 90.11 & 89.94 & 83.37 & 82.69 & 58.44 & 21.39 & \avgcell{70.99} \\
FlexPrefill-Qwen & \dash & 1.0
& 89.92 & 88.49 & 81.73 & 81.20 & 59.14 & 24.03 & \avgcell{70.75} \\

\bottomrule
\end{tabular}}
\end{table*}

\begin{table*}[h]
\centering
\caption{
Full InfiniteBench probe ablation.
Clip $p$ denotes norm clipping at percentile $p$, and Norm $\alpha$ denotes soft norm balancing.
The case $p=\text{--}$ and $\alpha=\text{--}$ is raw MASA without norm control.
\textbf{Bold} Avg. marks the best MASA variant within each setting.
}
\label{tab:app_infinitebench_probe_ablation}
\scriptsize
\setlength{\tabcolsep}{2.5pt}
\renewcommand{\arraystretch}{1.06}
\resizebox{\textwidth}{!}{
\begin{tabular}{llcccccccccccc}
\toprule
Setting & Clip $p$ & Norm $\alpha$
& En.Sum & En.QA & En.MC & En.Dia & Zh.QA
& \makecell[c]{Code\\Debug}
& \makecell[c]{Math\\Find}
& \makecell[c]{Retr.\\PassKey}
& \makecell[c]{Retr.\\Number}
& \makecell[c]{Retr.\\KV}
& \avgcell{Avg.} \\
\midrule

MInference & \dash & \dash
& 20.51 & 13.14 & 66.38 & 9.00 & 12.34
& 22.34 & 32.57 & 100.00 & 100.00 & 14.20
& \avgcell{39.05} \\
MInference & 90 & 0.6
& 21.13 & 13.02 & 66.38 & 8.00 & 12.31
& 23.10 & 32.00 & 100.00 & 100.00 & 14.80
& \avgcell{39.07} \\
MInference & 80 & 0.7
& 19.73 & 12.15 & 66.81 & 8.50 & 12.28
& 22.84 & 33.43 & 100.00 & 100.00 & 16.20
& \avgcell{\best{39.20}} \\

\midrule

FlexPrefill-Llama & \dash & 0.5
& 32.10 & 26.25 & 66.81 & 19.50 & 33.66
& 20.05 & 29.14 & 98.31 & 99.66 & 54.40
& \avgcell{\best{47.99}} \\

\midrule

FlexPrefill-Qwen & \dash & \dash
& 15.55 & 6.38 & 32.31 & 16.50 & 13.09
& 21.07 & 30.29 & 94.41 & 74.58 & 0.00
& \avgcell{\best{30.42}} \\
FlexPrefill-Qwen & \dash & 0.5
& 15.65 & 5.98 & 34.50 & 14.50 & 11.77
& 19.29 & 33.43 & 94.07 & 73.90 & 0.00
& \avgcell{30.31} \\

\bottomrule
\end{tabular}}
\end{table*}

\begin{table}[h]
\centering
\caption{
Full LongBench probe ablation for SeerAttention.
Clip $p$ denotes norm clipping at percentile $p$.
The case $p=\text{--}$ and $\alpha=\text{--}$ is raw MASA without norm control.
\textbf{Bold} Avg. marks the best MASA variant.
}
\label{tab:app_longbench_probe_ablation}
\small
\setlength{\tabcolsep}{5.0pt}
\renewcommand{\arraystretch}{1.08}
\resizebox{0.5\textwidth}{!}{
\begin{tabular}{llccc|c}
\toprule
Clip $p$ & Norm $\alpha$ & 0--4K & 4--8K & 8K+ & \avgcell{Avg.} \\
\midrule

\dash & \dash
& 55.31 & 54.57 & 52.84 & \avgcell{54.24} \\
85 & \dash
& 55.27 & 54.79 & 52.70 & \avgcell{54.25} \\
87.5 & \dash
& 55.45 & 54.48 & 53.16 & \avgcell{\best{54.37}} \\
90 & \dash
& 55.27 & 54.65 & 52.92 & \avgcell{54.28} \\

\bottomrule
\end{tabular}}
\end{table}

\end{document}